\documentclass[fleqn,10pt]{wlscirep}
\usepackage[utf8]{inputenc}
\usepackage[T1]{fontenc}
\usepackage{graphicx}
\usepackage{authblk}
\usepackage{siunitx}
\usepackage{booktabs}
\usepackage{caption}
\usepackage{float}
\usepackage{color,soul}
\usepackage{bm}
\usepackage{makecell}
\usepackage{xurl}
\usepackage{tabularx}   
\usepackage{multirow}
\usepackage{times}
\usepackage[title]{appendix}
\usepackage{amsmath}
\usepackage{amsfonts}
\usepackage{amssymb}
\usepackage{xcolor}
\usepackage{xr}

\usepackage{array}
\usepackage[super,sort&compress,numbers]{natbib}
\usepackage{hyperref}

\makeatletter

\newcommand*{\addFileDependency}[1]{% argument=file name and extension
\typeout{(#1)}% latexmk will find this if $recorder=0
\@addtofilelist{#1}
\IfFileExists{#1}{}{\typeout{No file #1.}}
}\makeatother

\newcommand*{\myexternaldocument}[1]{%
\externaldocument{#1}%
\addFileDependency{#1.tex}%
\addFileDependency{#1.aux}%
}
\myexternaldocument{supplementary}
\title{Retrofitting Earth System Models with Cadence-Limited Neural Operator Updates}
\author[1$\dagger$]{Aniruddha Bora}
\author[2$\dagger$]{Shixuan Zhang}
\author[1]{Khemraj Shukla}
\author[2]{Bryce E. Harrop}
\author[1$\ast$]{George Karniadakis}
\author[2$\ast$]{L. Ruby Leung}

\affil[1]{Division of Applied Mathematics, Brown University, RI, USA}
\affil[2]{Atmospheric, Climate, and Earth Sciences Division, Pacific Northwest National Laboratory, Richland, WA, USA}
\affil[$\dagger$]{Equal contributor}
\affil[$\ast$]{Corresponding author}
\begin{abstract}
Coarse resolution, imperfect parameterizations, and uncertain initial states and forcings limit Earth-system model (ESM) predictions. Traditional bias correction via data assimilation improves constrained simulations but offers limited benefit once models run freely. We introduce an operator-learning framework that maps instantaneous model states to bias-correction tendencies and applies them online during integration. Building on a U-Net backbone, we develop two operator architectures Inception U-Net (IUNet) and a multi-scale network (M\&M) that combine diverse upsampling and receptive fields to capture multiscale nonlinear features under Energy Exascale Earth System Model (E3SM) runtime constraints. Trained on two years E3SM simulations nudged toward ERA5 reanalysis, the operators generalize across height levels and seasons. Both architectures outperform standard U-Net baselines in offline tests, indicating that functional richness rather than parameter count drives performance. In online hybrid E3SM runs, M\&M delivers the most consistent bias reductions across variables and vertical levels. The ML-augmented configurations remain stable and computationally feasible in multi-year simulations, providing a practical pathway for scalable hybrid modeling. Our framework emphasizes long-term stability, portability, and cadence-limited updates, demonstrating the utility of expressive ML operators for learning structured, cross-scale relationships and retrofitting legacy ESMs.

\end{abstract}

\begin{document}

\flushbottom
\maketitle

\thispagestyle{empty}

\setcounter{secnumdepth}{0} 

\section{INTRODUCTION}
Machine learning (ML) has emerged as a powerful tool for scientific computing, offering flexible function approximation, operator learning, and multi-fidelity data fusion. ML can efficiently capture nonlinear relationships, emulate complex dynamics, and integrate information across scales, making it particularly well suited to problems where high-fidelity simulations are scarce but lower-fidelity data are abundant. This  capability aligns directly with the challenges of Earth system modeling, where computational resource limits necessitate coarse-resolution simulations that suffer from systematic biases, while high-resolution models remain prohibitively expensive for long-range scientific simulations. Harnessing the synergy between Earth system modeling and ML has therefore become increasingly important, enabling the development of hybrid approaches that couple physical understanding with data-driven efficiency.

One promising pathway is the design of hybrid frameworks that embed ML operators within Earth system models. Such integration can preserve the physical fidelity of ML-based predictions while improving simulation skill. For example, Watt-Meyer et al. \cite{watt2021correcting} implemented an online bias-correction scheme in which a random forest model was trained on nudged historical simulations toward a global reanalysis to predict the tendency corrections required to reduce systematic errors in a global atmospheric model. Although this approach reduced certain mean-state biases, tree-based methods such as random forests are inherently limited as they operate as local, point-wise predictors that are ill-suited to capturing spatially and temporally coherent structures, they offer restricted flexibility for representing nonlinear multiscale interactions and generalize poorly when applied beyond the training distribution. In parallel, end-to-end differentiable hybrids such as NeuralGCM train learned parameterizations jointly with a differentiable dynamical core, optimizing directly to observations for weather-through-climate skill \cite{kochkov2024neural}. Our plug-in neural operator corrector has a different aim and deployment pathway: it retrofits a legacy GCM to deliver low-overhead, cadence-limited tendency updates that prioritize long-term online stability and portability without end-to-end retraining. These considerations highlight the need for more expressive ML frameworks capable of learning complex but structured relationships across scales.

Recent advances in operator learning provide suitable potential frameworks. Rather than approximating finite-dimensional inputs and outputs, operator learning seeks to learn mappings between functional spaces \cite{lu2021learning,li2020fourier,kovachki2021neural,li2020multipole}. This perspective has already led to significant breakthroughs in scientific machine learning, including advances in weather forecasting at unprecedented resolution and lead times \cite{lam2023learning,bi2023accurate,kurth2023fourcastnet}, suggesting that similar approaches could offer a principled pathway for enhancing bias correction in climate models.

In this work, we propose two new architectures for online bias correction of low-resolution climate models that leverage multiple latent spaces to improve representation learning. Both architectures are based on UNets \cite{cciccek20163d}, but we demonstrate that, given the same number of trainable parameters, more expressive decoder designs substantially enhance feature extraction and predictive skill. We have successfully integrated these trained machine learning frameworks with the Energy Exascale Earth System Model (E3SM) \cite{Golaz2022E3SMv2} to perform online bias correction and demonstrated the viability and promise of hybrid neural operators-physics based climate modeling for improving climate simulations and predictions at low resolution. 

\section{RESULTS}
\begin{figure}[!ht]
    \centering
    \includegraphics[width=0.9\linewidth]{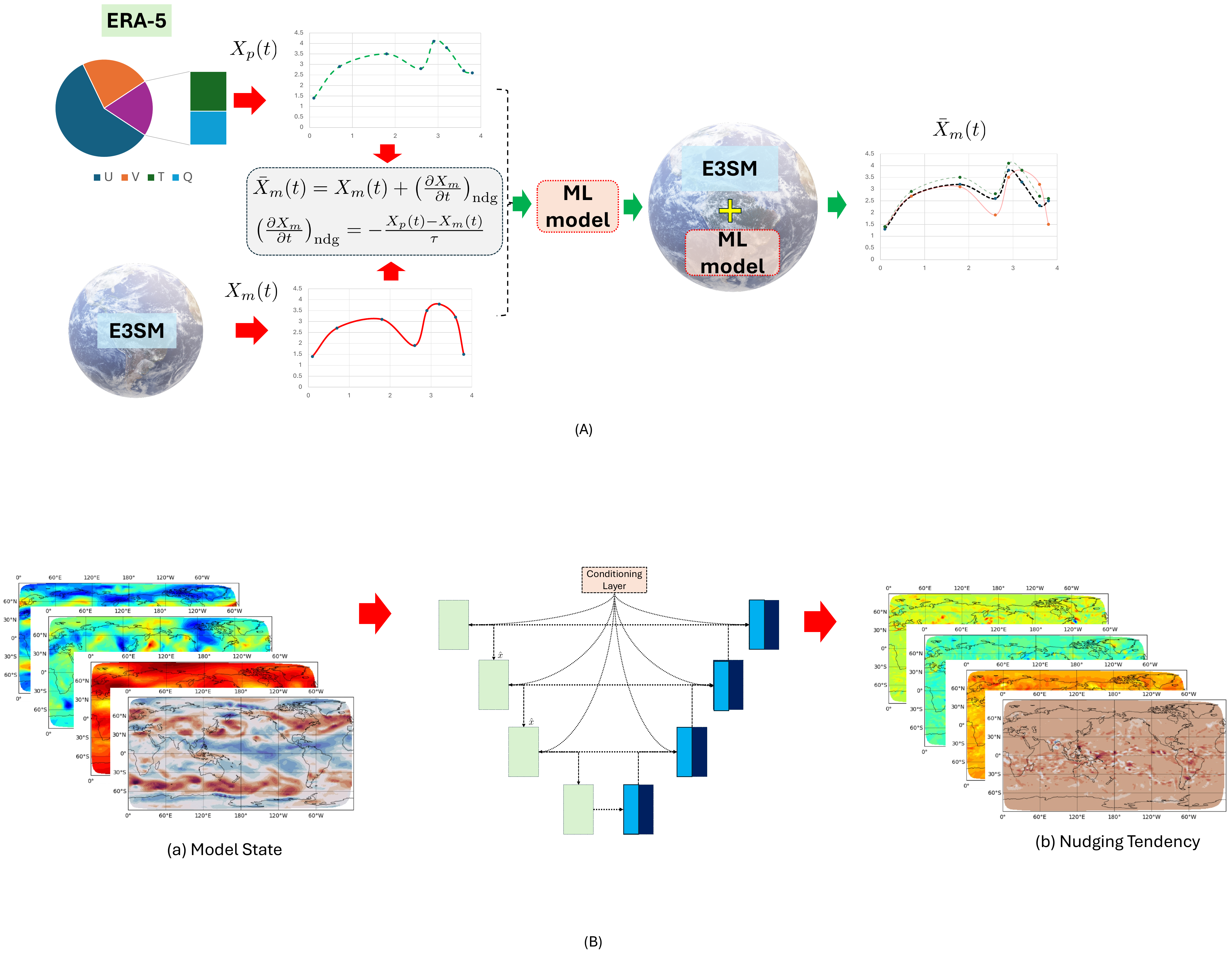}
    \caption{\textbf{Workflow and learning task (M\&M with FiLM conditioning).} \textbf{(A)} \emph{End-to-end workflow.} ERA5 provides the reference data and the U.S. Department of Energy’s Energy Exascale Earth System Model (E3SM) provides simulated states. Controlled (nudged) E3SM runs toward ERA5 yield paired \emph{model states} and \emph{nudging tendencies} used to train the ML operator, which is then coupled back into E3SM for online bias correction during free-running simulations (no reference data). \textbf{(B)} \emph{Inputs and targets.} \textbf{(a)} Pre-nudging state variables zonal wind $U$, meridional wind $V$, temperature $T$, and specific humidity $Q$ serve as inputs. \textbf{(b)} The learned model maps these states to the corresponding nudging tendencies. The central network schematic in (B) depicts the \emph{M\&M} architecture with \emph{FiLM} scalar conditioning; full architectural details are provided in the Methods, and expanded layer-level diagrams appear in the Supplementary Information (see Supplementary Figures~S14 and~S15).}
    \label{fig:1}
\end{figure}

The framework of our ML bias correction is described in Figure~\ref{fig:1} (panel A), along with the inferred best architecture (panel B). Here, we cast bias correction as a discrete neural-operator map from the instantaneous E3SM Atmosphere Model (EAM) simulated state \(X_m\) to a nudging tendency \(\left(\partial X_m / \partial t\right)_{\text{ndg}}\) that updates the model in a single step, so that the corrected state \(\bar{X}_m = X_m + \left(\partial X_m / \partial t\right)_{\text{ndg}}\) tracks ERA5 (Eq.~\eqref{eq3}). Two implementation constraints shape the design: online E3SM exposes only the modeled state from the immediate previous timestep (no history) and runs on CPUs, motivating an image-to-image formulation with per-channel normalization and a compact UNet backbone. Building on this, we introduce \textbf{IUNet} (multi-path Inception blocks with learned metadata/scalar embeddings injected at every layer) and \textbf{M\&M} (a full-rank, artifact-resistant, three-branch upsampling decoder). Trained on 2009--2010 and evaluated out-of-year on 2015, the offline results show that conditioning and decoder design are decisive: a $\sim$0.2~M-parameter UNet gains sharply with a simple scalar embedding (pattern correlation from 0.22/0.26 for U/V to 0.75/0.72), and at matched capacity ($\sim$6.7~M) M\&M attains the lowest errors (RMSE/MAE), highest PSNR/SSIM, and global \(R^2\) up to 0.281 (U) and 0.206 (V), with temporal correlations remaining positive throughout most of the atmospheric column and strongest near the surface (Figs.~\ref{fig:1}, 4--6; note that larger vertical-layer indices correspond to layers closer to the surface). We then implement the leading candidates online in EAM: ML-corrected runs reproduce the spatial structure of vertically averaged tendencies (UTEND/VTEND/TTEND/QTEND) with \(\mathrm{PCC} \approx 0.7\text{--}0.85\), reduce seasonal RMSE at pressure- and surface-level variables, weaken zonal-wind biases throughout the column, and damp 200-hPa temperature biases over storm tracks (Figs.~7--10). In the sections that follow, we detail offline architecture/ablations, skill, and online coupling diagnostics and limitations.

\subsection{Effect of scalar embedding (FiLM conditioning)}
Scalar (FiLM)\cite{perez2018film} embedding substantially improves prediction skill across all UNet-based architectures (see Supplementary Figure~S1). In a $\sim$0.2M parameter UNet trained without conditioning, the mean nudging tendency for the test year shows large-scale spatial errors, notably over Antarctica and central Asia (see Supplementary Figures~S1 and ~S14). Augmenting the same UNet with FiLM-based scalar encoding markedly increases spatial pattern agreement—raising the mean pattern correlation by up to $\sim$3$\times$ and reducing regional biases. Motivated by these gains, all results reported henceforth use FiLM-conditioned variants for each architecture.

 \subsection{\textbf{Offline testing and verification}}\label{sec:offline_results}
\begin{figure}[!ht]
\centering
\includegraphics[width=\linewidth]{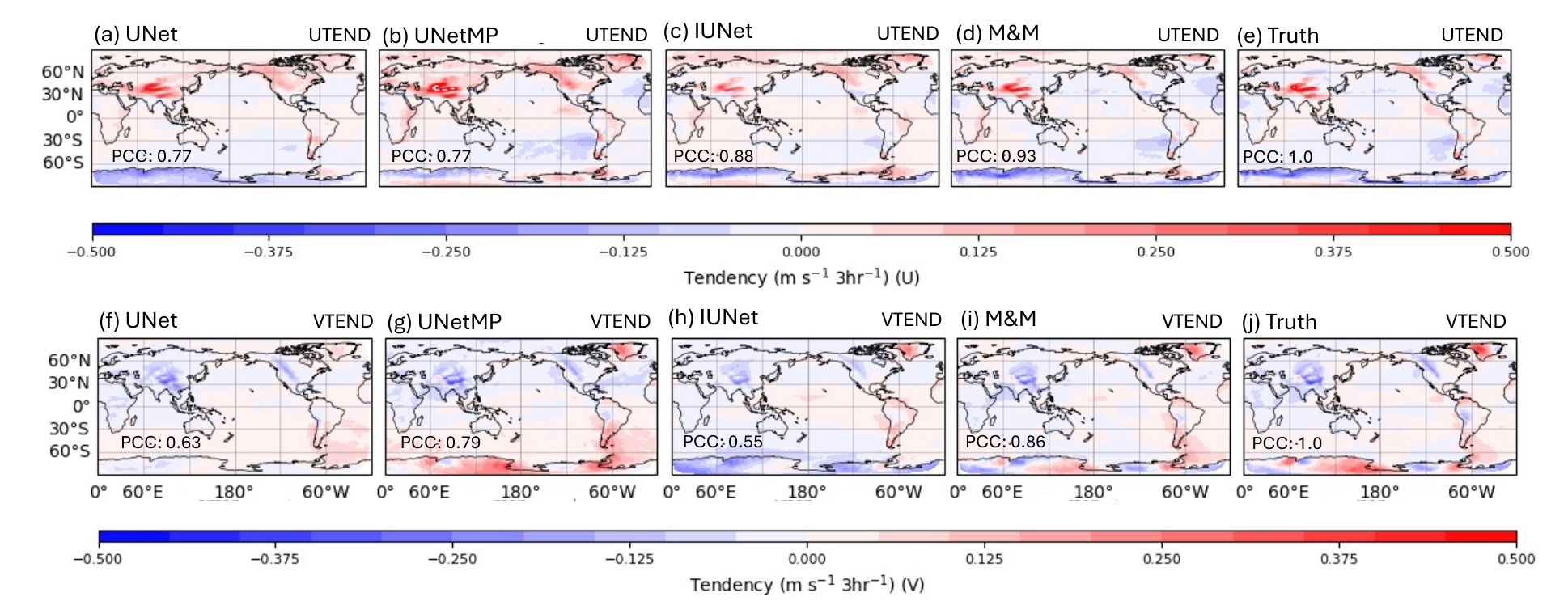}
\caption{\textbf{Offline comparison of mean wind nudging tendencies across neural-operator architectures.} Panels (a–e) show zonal wind correction tendencies (UTEND) and panels (f–j) show meridional wind correction tendencies (VTEND), each averaged over all vertical levels and over the year 2015. For each component, global maps from UNet, UNet with increased parameters (UNetMP), IUNet, and the multi-branch full-rank decoder (M\&M) are shown alongside the reference nudging tendencies from the training data (“Truth”). Values within each panel denote the spatial pattern correlation (PCC, Pearson correlation) with the reference field, summarizing large-scale spatial agreement. Across both UTEND and VTEND, M\&M exhibits the strongest correspondence with the reference tendencies, followed by UNetMP, while UNet and IUNet show more mixed performance. All models were trained on the same dataset with identical batch size, optimizer, and number of epochs to ensure a controlled comparison.}
\label{fig:UNdg_mean}
\end{figure}

\begin{figure}[!ht]
\centering
\includegraphics[width=\linewidth]{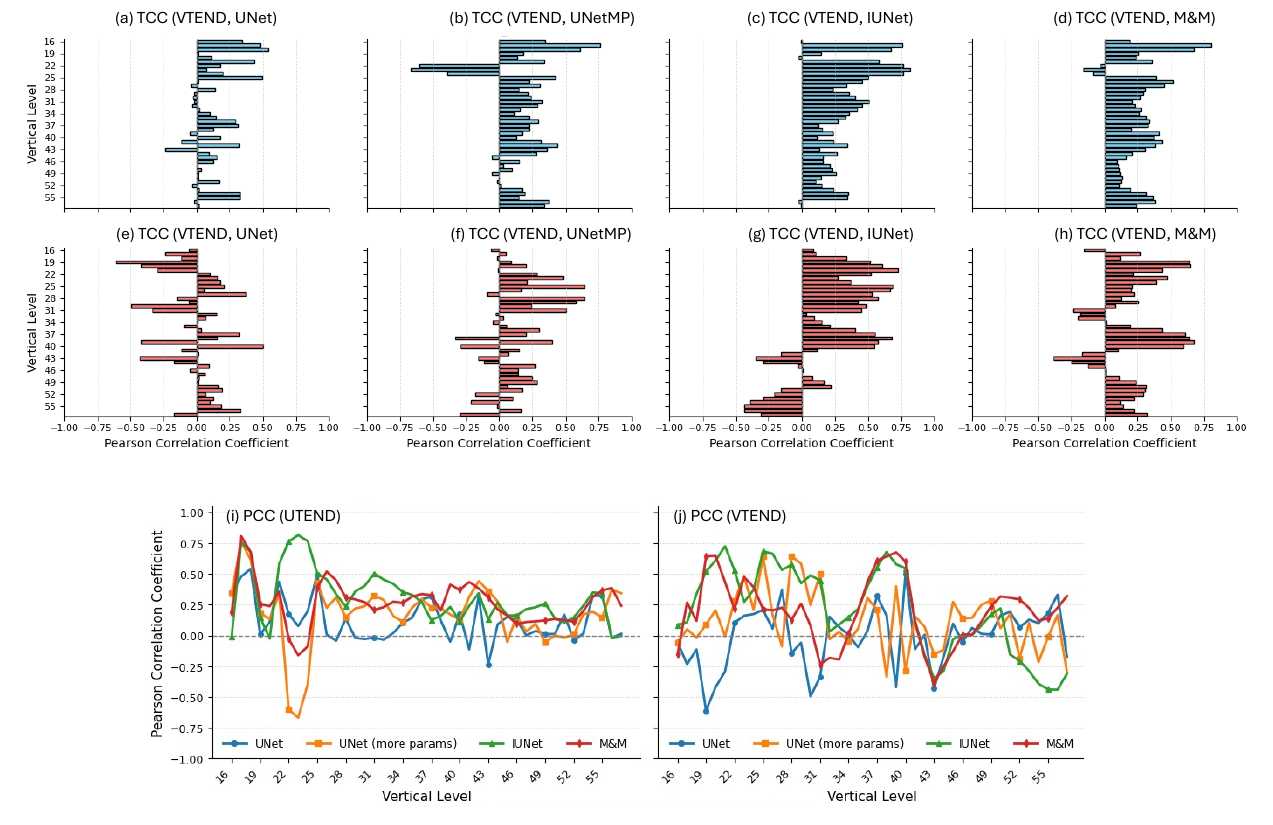}
\caption{\textbf{Offline comparison of vertical-layer skill for wind nudging tendencies across neural-operator architectures.} Panels (a–h) show layer-wise \emph{temporal correlation coefficients} (TCC; Pearson correlation over time at each vertical level) between predicted and reference nudging tendencies (training datasets) for zonal (UTEND; top row) and meridional (VTEND; bottom row) wind components. Columns correspond to different architectures: UNet, UNet with increased parameters (UNetMP), IUNet, and the multi-branch full-rank decoder (M\&M). Positive correlations indicate better temporal tracking of reference tendencies, whereas negative values reflect phase mismatch. The vertical-level index increases toward the surface (higher index = nearer-surface levels). Panels (i–j) show the \emph{spatial pattern correlation} at each level, computed as the Pearson spatial correlation between predicted and reference fields at every time step and then averaged over 2015. Curves summarize this mean pattern correlation for UTEND (i) and VTEND (j), using the same vertical-level indexing as in panels (a–h). Higher values indicate better reproduction of the spatial structure at each level.}
\label{fig:UNdg_mean_TC}
\end{figure}

For offline testing, we compare the different methodologies over the test year 2015, where the models were trained with data for the year 2009-2010. To assess each architecture's spatiotemporal fidelity for the test year 2015, we show plots for these complementary diagnostics: (a) bulk temporal correlation, which is the Pearson correlation between the true and predicted nudging tendencies averaged over all time steps and vertical layers (see Figure \ref{fig:UNdg_mean}), (b) layer-wise time correlation, which is the Pearson correlation computed at each pressure level to reveal height dependent tracking (see Figure~\ref{fig:UNdg_mean_TC}). We focus on these mean metrics because we train our models to reproduce the mean nudging tendency as specified in our problem statement. When implemented in the online integration framework of E3SM, accurately matching this mean ensures the systematic errors are cumulatively removed over successive model time steps, driving improved performance in long-term simulations.   Moreover, we evaluate the different models across different error metrics such as MSE (mean squared error), RMSE (root mean squared error), MAE (maximum absolute error), PSNR (Peak Signal-to-Noise Ratio), Bias, StdError (Standard Error), Global $R^{2}$ (Coefficient of Determination), CV error (Coefficient of Variation of the Error) and SSIM (Structural Similarity Index ). These metrics are defined in Supplementary Section~S2. Even seemingly small improvements in these mean metrics matter; since the correction is applied at every model time step or once every 6 model steps, those incremental gains compound over time to yield a significant reduction in climate bias. Tables~\ref{tab:UVcomp}, Supplementary Figures~S2 and~S3 show the comparison for all the different models for the $U$(Zonal Wind), $V$(Meridional Wind), $Q$(Humidity), and $T$(Temperature), respectively. To ensure a fair comparison between architectures with similar parameter counts, all models were trained on the same dataset using the same batch size and optimization settings for an identical number of epochs, so that each model processed the same number of training examples and gradient updates.

Figure \ref{fig:UNdg_mean} (a) and Table~\ref{tab:UVcomp} summarize the performance of the four architectures in reproducing the global mean nudging tendency of U‑zonal wind (m/s) across all vertical layers and time for the test year 2015. The baseline UNet ($\sim$0.2M parameters) attains a pattern correlation of 0.77, which still remains almost same when scaled up to 6.7M parameters. IUNet (6.7M) improves upon it with a correlation of 0.88. M\&M  (6.7M) reaches the highest pattern correlation of 0.93, closely matching the spatial structure of the ground truth. Moreover, M\&M achieves the lowest MSE ($2.36\times 10^{-4}$), RMSE ($1.53\times 10^{-2}$), and MAE ($1.01\times 10^{-2}$), along with the highest PSNR (36.25) and SSIM (0.48). It also exhibits the smallest bias ($3.652\times 10^{-5}$), lowest standard error ($1.539\times 10^{-2}$), highest global R² (0.208), and lowest coefficient-of-variation error (0.0339). These results demonstrate that M\&M significantly outperforms both UNet and IUNet in capturing both global and local patterns of the nudging tendency, without increasing the parameter budget. \begin{table}[htbp]
\centering
\caption{\textbf{Offline performance comparison of neural network architectures.}
Global-mean metrics for nudging tendencies of the zonal wind ($U$) and meridional wind ($V$).
Best (or tied-best) values are shown in bold math.}
\label{tab:UVcomp}
\scriptsize
\begin{tabular}{llcccc}
\toprule
\textbf{Field} & \textbf{Metric} & \textbf{UNet} & \textbf{UNet (more params)} & \textbf{IUNet} & \textbf{M\&M} \\
\midrule
\multirow{10}{*}{$U$ (zonal)}
& Parameters            & $\mathbf{\sim 0.2\,\mathrm{M}}$ & $\sim 6.7\,\mathrm{M}$ & $\sim 6.7\,\mathrm{M}$ & $\sim 6.7\,\mathrm{M}$ \\
& MSE                   & $2.67550\times 10^{-4}$ & $2.47464\times 10^{-4}$ & $2.42019\times 10^{-4}$ & $\mathbf{2.36928\times 10^{-4}}$ \\
& RMSE                  & $1.63570\times 10^{-2}$ & $1.57310\times 10^{-2}$ & $1.55570\times 10^{-2}$ & $\mathbf{1.53925\times 10^{-2}}$ \\
& MAE                   & $1.07062\times 10^{-2}$ & $1.04996\times 10^{-2}$ & $1.02471\times 10^{-2}$ & $\mathbf{1.01759\times 10^{-2}}$ \\
& PSNR                  & $35.73$ & $36.06$ & $36.16$ & $\mathbf{36.25}$ \\
& Bias                  & $5.294\times 10^{-4}$ & $5.488\times 10^{-4}$ & $5.026\times 10^{-4}$ & $\mathbf{3.652\times 10^{-5}}$ \\
& StdError              & $1.635\times 10^{-2}$ & $1.572\times 10^{-2}$ & $1.555\times 10^{-2}$ & $\mathbf{1.539\times 10^{-2}}$ \\
& Global $R^2$          & $0.105$ & $0.172$ & $0.191$ & $\mathbf{0.208}$ \\
& CV error              & $0.0360$ & $0.0346$ & $0.0342$ & $\mathbf{0.0339}$ \\
& SSIM                  & $0.4526$ & $0.4683$ & $0.4762$ & $\mathbf{0.4872}$ \\
\midrule
\multirow{10}{*}{$V$ (meridional)}
& Parameters            & $\mathbf{\sim 0.2\,\mathrm{M}}$ & $\sim 6.7\,\mathrm{M}$ & $\sim 6.7\,\mathrm{M}$ & $\sim 6.7\,\mathrm{M}$ \\
& MSE                   & $2.56221\times 10^{-4}$ & $2.45302\times 10^{-4}$ & $2.41353\times 10^{-4}$ & $\mathbf{2.35770\times 10^{-4}}$ \\
& RMSE                  & $1.60069\times 10^{-2}$ & $1.56621\times 10^{-2}$ & $1.55355\times 10^{-2}$ & $\mathbf{1.53548\times 10^{-2}}$ \\
& MAE                   & $1.06612\times 10^{-2}$ & $1.06106\times 10^{-2}$ & $1.04069\times 10^{-2}$ & $\mathbf{1.03080\times 10^{-2}}$ \\
& PSNR                  & $35.91$ & $36.10$ & $36.17$ & $\mathbf{36.28}$ \\
& Bias                  & $3.130\times 10^{-5}$ & $-6.283\times 10^{-6}$ & $-8.888\times 10^{-5}$ & $\mathbf{-1.434\times 10^{-6}}$ \\
& StdError              & $1.601\times 10^{-2}$ & $1.566\times 10^{-2}$ & $1.554\times 10^{-2}$ & $\mathbf{1.535\times 10^{-2}}$ \\
& Global $R^2$          & $0.121$ & $0.158$ & $0.172$ & $\mathbf{0.191}$ \\
& CV error              & $0.0352$ & $0.0345$ & $0.0342$ & $\mathbf{0.0338}$ \\
& SSIM                  & $0.3936$ & $0.4034$ & $0.4072$ & $\mathbf{0.4217}$ \\
\bottomrule
\end{tabular}
\end{table}

%%%%%%%%%%%%

Figure \ref{fig:UNdg_mean} (b) and Table~\ref{tab:UVcomp} compare how each model reconstructs the global mean nudging trend for the meridional wind (V) in 2015, both in terms of spatial pattern and amplitude distribution. The lightweight UNet ($\sim$0.2M parameters) achieves a pattern correlation of 0.63 but compresses anomalies and does not reproduce peaks over the Southern Ocean or strong north lobes in mid-latitudes. Scaling the UNet to 6.7M parameters raises its correlation to 0.79 and modestly widens the anomaly range, yet extremes remain systematically underestimated. IUNet (6.7M) achieved a pattern correlation of 0.55 and produced an enhanced correction tendency over tropical regions compared to other architectures, while still attenuating high-latitude extrema, as indicated by its more subdued blue tail. In contrast, M\&M (6.7M) matches the ground truth’s distribution while attaining a pattern correlation of 0.86. In contrast, M\&M (6.7M) matches the ground truth’s distribution while attaining a pattern correlation of 0.78. Quantitatively, M\&M deliver the lowest MSE ($2.35\times 10^{-4}$), RMSE ($1.53\times 10^{-2}$) and MAE ($1.03\times 10^{-2}$), the highest PSNR (36.28) and SSIM (0.42), near‑zero bias ($-1.434\times10^{-6}$), the smallest standard error ($1.53\times 10^{-2}$), the greatest global R² (0.191) and the lowest coefficient‑of‑variation error (0.0338). Figure \ref{fig:UNdg_mean_TC} (a) shows the temporal fidelity of the models for zonal wind U for the vertical levels between 14 and 58 for the test year 2015. The UNet with fewer parameters $\sim$0.2M achieves correlation of the order $\sim$0.3-0.5 in some regions of the mid tropospheric levels, but is near zero almost everywhere else. Scaling the UNet to $\sim$6.7M parameters increases the correlation to $\sim$0.75 at around 18 out of the 44 levels. IUNet with the same number of parameters improves the temporal coherence across all the levels with correlation values of $\sim$0.6 - 0.8 for most of the levels. M\&M with the same number of parameters delivers the most consistent time correlations. Figure \ref{fig:UNdg_mean_TC} (b) shows the temporal fidelity of the models for meridional wind V for the vertical levels between 14 to 58 for the test year 2015. The UNet with less parameters shows erratic behavior. It has positive correlation values of the range $\sim$0.3-0.4 for levels between 26-28 and above level 55, but it goes to  $\sim$-0.6 at mid tropospheric layers, which indicates severe misalignment. Scaling the UNet to $\sim$6.7M parameters increases the positive correlation at level 26-29 and reduces the negative magnitude of the correlations, but more or less the oscillations in the correlation value still exist. IUNet with the same number of parameters increases correlation sustaining correlations above 0.5 from levels 19 through 34, but still suffers pronounced negative correlations in the near surface layers. The M\&M architecture, on the other hand, with the same number of parameters, delivers the most consistent and highest temporal fidelity. Its peak value is $\sim$0.75 at the lower troposphere, and maintains a positive correlation above 0.3 from the near surface to level 40 and confines negative correlation only to a few levels.
Figure \ref{fig:UNdg_mean_TC} (c) plots the layer-wise mean Pearson correlation over all time for the test year 2015. The UNet with less parameters achieves a weak correlation of around 0.2 over much of the column and negative values near the surface and upper troposphere. Scaling the UNet to $\sim$6.7M parameters lifts the mid tropospheric correlations to the order of 0.4-0.6 and reduces the negative correlation values over the upper layers, but still exhibits oscillations. IUNet with the same number of parameters further increases the positive correlation values up to 0.6 and above for most of the levels, but collapses to negative in the lowermost layers. M\&M achieves the highest and the most uniform mean correlations. It peaks around 0.7-0.75 in the upper troposphere, remains above 0.4 through level 35, and only gently tapers towards zero near the surface layer. 
Across both zonal (U) and meridional (V) winds, UNet with fewer parameters struggles to capture the full spatial variability, yielding low pattern correlations throughout. Increasing the parameters of UNet to $\sim$6.7 M further sharpens the spatial characteristics and increases the temporal coherence, but still underestimates and exhibits negative correlation at several levels. IUNet with the same number of parameters enhances the predictions further and increases the correlations for most upper and middle layers, but still has degraded performance over the lower troposphere. M\&M with the same number of parameters consistently achieves the highest correlation and sustains strong temporal coherence with slight degradation towards the bottom of the column, demonstrating superior spatial and temporal fidelity for both U and V.

\subsubsection{Online Implementation strategy}
We employ the officially released version 2 of the E3SM atmosphere model (EAMv2) \cite{Golaz2022E3SMv2} to conduct online tests of the machine learning (ML) approaches discussed in the previous section. The ML model is integrated into EAMv2 as a modular component, conceptually similar to the nudging implementation described in~\nameref{sec:training_data}. However, Eq.~(\ref{eq1}) is reformulated as:
\begin{equation} \label{eqn:ml_nudging}
    \frac{\partial X_m}{\partial t} = D(X_m) + F(X_m) + \mathbb{M}(X_m; t),
\end{equation}
where the left-hand side and the first two terms on the right-hand side retain the same meaning as in Eq.~(\ref{eq1}). The third term on the right, $\mathbb{M}(X_m; t)$, represents the ML-predicted correction, which replaces the term of the nudge tendency in Eq.~(\ref{eq1}).

As discussed in the previous section, the original nudging tendency was computed as a 3-hourly average to suppress high-frequency noise. Consequently, the ML training was formulated to predict the 3-hour mean nudging tendency using the instantaneous model state at the beginning of each window. To remain consistent with this setup, the online application of the ML correction was implemented as:
\begin{equation} \label{eqn:ml_form}
    \mathbb{M}(X_m; t) = 
    \begin{cases}
        \mathbb{M}(X_m), & \text{if } t = t_0, \\
        0, & \text{otherwise,}
    \end{cases}
\end{equation}
In this formulation, \( t_0 \) denotes the start time of each 3-hour window. The ML correction is applied as an \textit{instantaneous tendency} at \( t_0 \), and is set to zero at all other times within the window. This approach ensures consistency with the offline training setup, in which instantaneous model states were used at the beginning of each 3-hour window to predict the corresponding 3-hour mean nudging tendencies. In principle, \( t_0 \) can be aligned with the EAMv2’s native time step (e.g., every 30 minutes), allowing for more frequent ML updates. However, short-term test simulations conducted during the initial evaluation of the online implementation (not shown) indicated that such frequent invocation of the ML model not only leads to a significant increase in computational cost but also disrupts the model’s internal dynamics and physical parameterizations. This disruption arises from a fundamental limitation of the offline training paradigm: the ML model is trained without exposure to the interactions between the correction tendencies and the model’s own dynamical and physical processes or feedback. Consequently, a more robust implementation strategy is needed to balance computational efficiency and correction performance. A systematic investigation of such a strategy is beyond the scope of this study and will be presented in a separate work.

In addition, the coupling of $M$, the PyTorch-based trained model, with FORTRAN was achieved by further developing the pytorch-fortran package, originally developed by Dmitry Alexeev et al.~\cite{alexeedm}. This package provides an interface to load the PyTorch traced model in the FORTRAN programming language, which is used to develop E3SM. It enables developers working with legacy or high-performance scientific computing codes in Fortran to directly invoke models exported in TorchScript format via a FORTRAN proxy library. Additional implementation details are provided in Supplementary Section~S3. A series of EAMv2 simulations was conducted with online bias corrections implemented using the UNet, UNetMP, IUNet, and M\&M architectures, as described in~\nameref{sec:ml_methodology}. All simulations employed the same model configuration used to generate the ML training data described in~\nameref{sec:training_data}. In addition, a free-running EAMv2 control simulation (CLIM) was performed to serve as the baseline for evaluating ML-based bias correction. Each simulation was integrated over the five-year period 2012–2016. Importantly, the evaluation years (2012–2016) were excluded from the offline training period (2009–2010), ensuring a fully independent validation. All ML bias-correction simulations remained numerically stable throughout integration, a robustness attributed to the implementation strategy in which corrections are applied as a gentle additional tendency term that minimally perturbs the native dynamics and physical processes of the model. A concise summary of these simulations is provided in Supplementary Table~S5.

\subsection{Online performance of ML bias corrections} \label{sec:online_perofrmance}
\begin{figure}[ht]
    \centering
    \includegraphics[width=\linewidth]{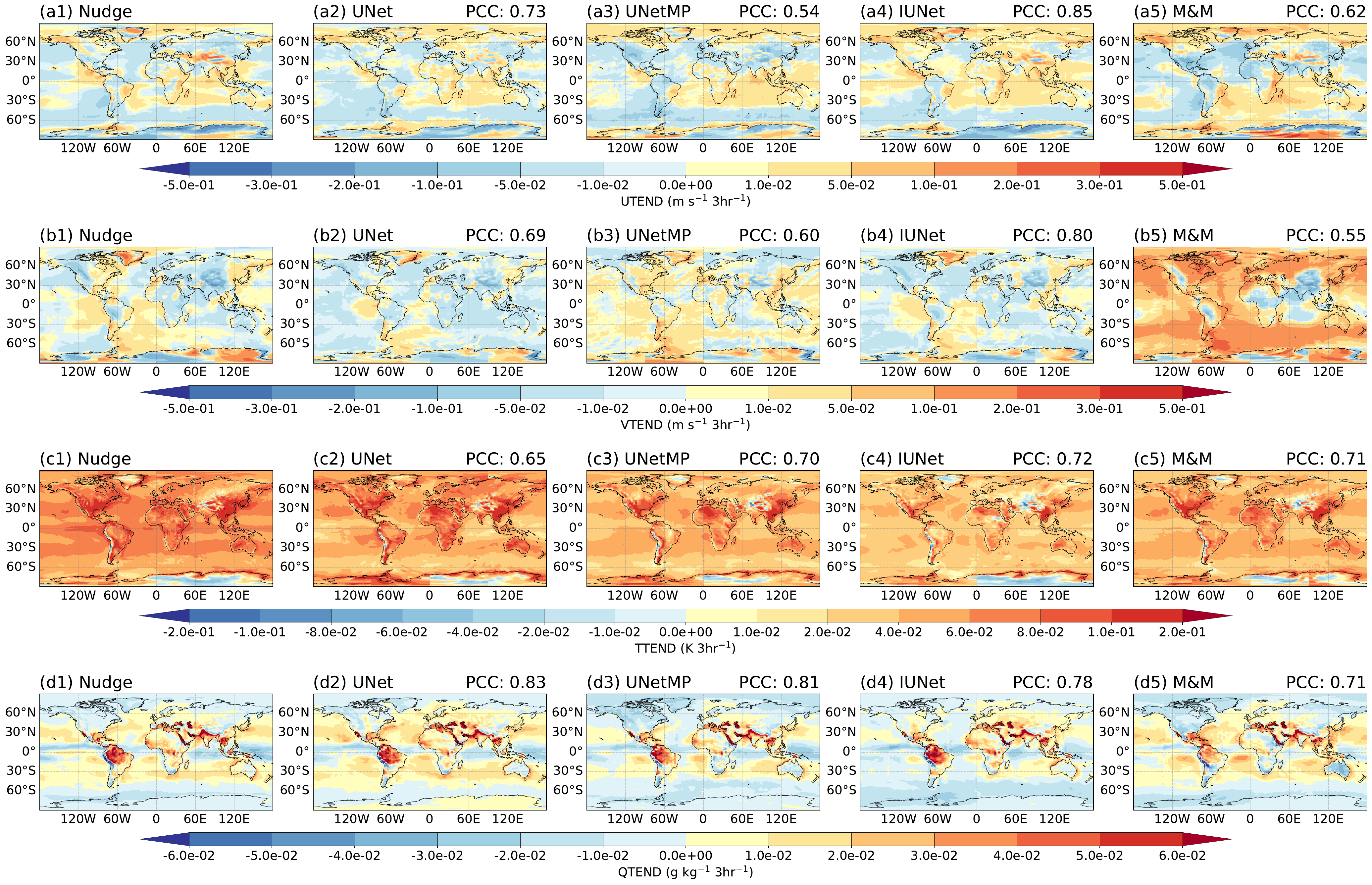}
    \caption{\textbf{Global distributions of annual-mean nudging tendencies.} 
    Zonal wind (UTEND, m\,s$^{-1}$\,3\,hr$^{-1}$; panels a1--a5), meridional wind (VTEND, m\,s$^{-1}$\,3\,hr$^{-1}$; panels b1--b5), air temperature (TTEND, K\,3\,hr$^{-1}$; panels c1--c5), and specific humidity (QTEND, g\,kg$^{-1}$\,3\,hr$^{-1}$; panels d1--d5) are averaged over all model levels and the 5 years of 2012--2016. The first column shows nudging tendencies from the reference simulation (Nudge), which serves as the target for model comparison. Columns two through five show results from different model configurations, including the free-running baseline (CTRL) and machine learning--corrected simulations (UNet, UNetMP, IUNet, and M\&M), as summarized in Supplementary Table~S5. Color contours represent the magnitude and sign of the vertically averaged nudging tendencies, with cool colors indicating negative values and warm colors positive values. Pattern Correlation Coefficients (PCCs) relative to the reference are shown in the upper right corner of each panel to quantify spatial agreement. All tendencies are scaled to 3-hourly values for consistency. Details of the simulation configurations are provided in Supplementary Table~S5.}
    \label{fig:ann_nudge_tend}
\end{figure}

Figure~\ref{fig:ann_nudge_tend} compares annual-mean correction tendencies from the ML models with the reference nudging simulation. All models reproduce the main spatial structures, with comparable skill in terms of pattern correlation coefficient (PCC) for temperature (0.65--0.72) and humidity (0.71--0.83), consistent with the offline evaluation in the previous section. UNet and M\&M also show stronger agreement in the sign and magnitude of tendencies over midlatitude storm-track regions compared to UNetMP and IUNet. By contrast, performance for winds ($U$ and $V$ components) is more variable: IUNet achieves the best agreement (PCC of 0.8--0.85), while M\&M and UNetMP perform less favorably (PCC of 0.5--0.6), with M\&M notably overcorrecting $V$ winds across most regions. Similar conclusions hold for the seasonal means (see Supplementary Figure~S9 for an example). The weaker performance of UNetMP relative to UNet suggests that increased parameterization does not necessarily enhance online skill; instead, it can promote overfitting and reduce robustness. As discussed in~\nameref{sec:offline_results}, this issue is particularly evident for variables with lower variance, such as temperature, where highly parameterized models may capture small fluctuations or noise rather than the underlying stable signal. In online simulations, such overfitting can be further exacerbated by nonlinear interactions, whereby errors in one variable (e.g., temperature) propagate to others and amplify overall biases. Overall, the ML models capture thermodynamic corrections reliably, whereas wind corrections remain more difficult, likely owing to their intrinsic high spatiotemporal variability.

\begin{figure}[ht]
    \centering
    \includegraphics[width=\linewidth]{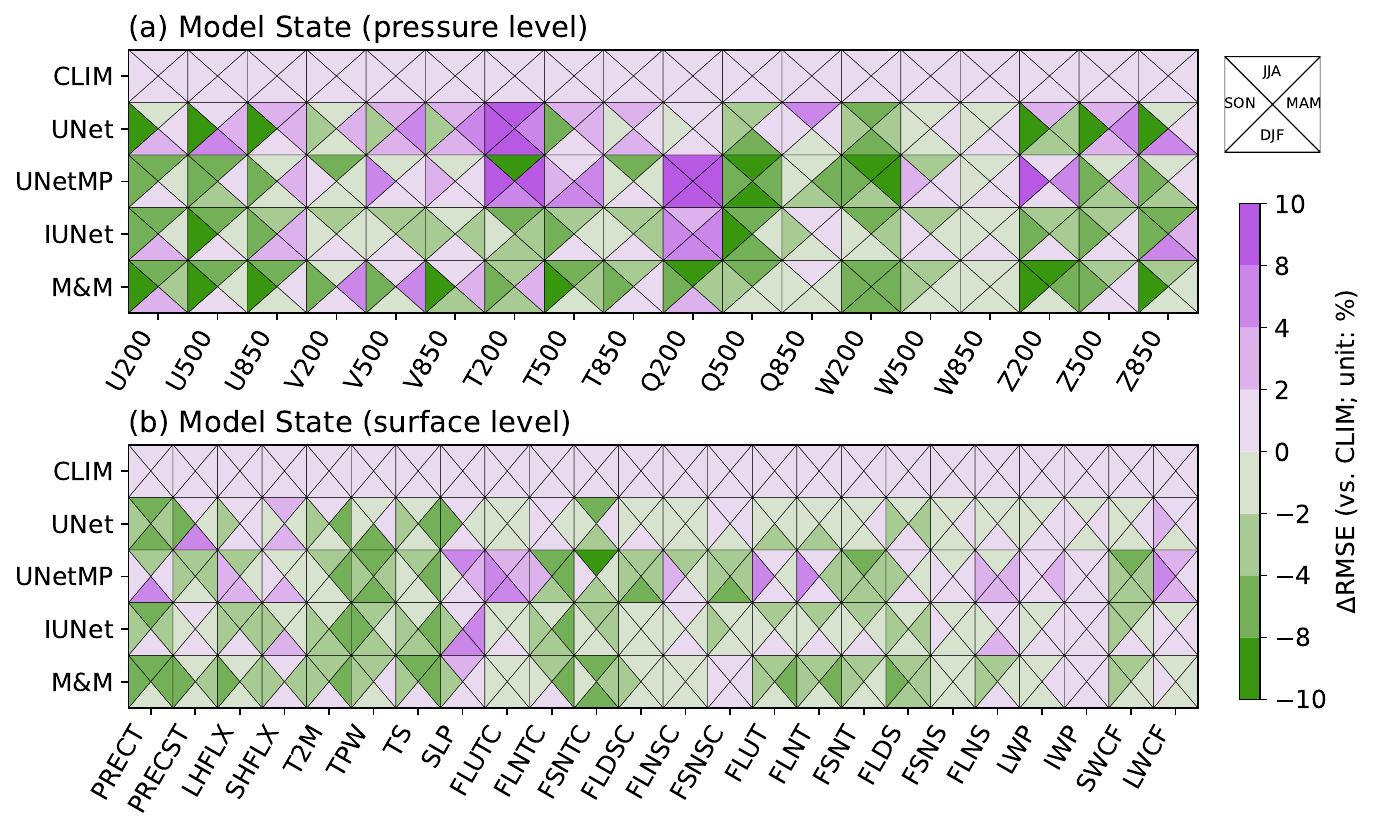}
    \caption{\textbf{Seasonal root-mean-square error (RMSE) averaged over 2012--2016.} Shown are the atmospheric state variables at pressure levels (top) and surface levels (bottom). Each square is divided into four triangles representing DJF (December--February), MAM (March--May), JJA (June--August), and SON (September--November). Colors indicate the percent difference in RMSE relative to the free-running EAMv2 baseline simulation (CLIM), with negative values indicating a reduction (improvement) and positive values an increase (degradation). Rows correspond to experiments and columns to variables (labels follow standard CMIP/CF aliases). Details of the simulation configurations are provided in Supplementary Table~S5.}
    \label{fig:rmse_heatmap_5yr}
\end{figure}

Following the nudging tendency evaluation, we assessed the overall impact of online ML bias correction by analyzing global-mean root-mean-square errors (RMSEs) in key dynamical and physical quantities at both upper-atmosphere and surface levels, as simulated by the EAMv2 model with and without ML-based corrections. As shown in Figure~\ref{fig:rmse_heatmap_5yr}, the ML models---particularly IUNet and M\&M---achieved a 2--10\% reduction in RMSE for most quantities and seasons relative to the baseline simulation (CLIM; first row). In contrast, the UNet-based architectures (UNet and UNetMP) led to substantial degradations in temperature and humidity, especially in the upper troposphere, consistent with their inability to reproduce the large-scale structures of the correction tendencies (Figure~\ref{fig:ann_nudge_tend}a2--b2, a3--b3). The performance of IUNet and M\&M also exhibited strong seasonal dependence, with all models struggling to reduce errors during boreal winter (DJF). Between the two, M\&M generally outperformed IUNet by producing greater RMSE reductions across several quantities. However, it showed limited skill in correcting the meridional wind ($V$), consistent with its tendency to overcorrect the nudging tendencies of this field (Figure~\ref{fig:ann_nudge_tend}b5). We further note that in the related ML bias correction framework of Watt-Meyer et al.(2021)$^{\cite{watt2021correcting}}$ for the FV3GFS model, precipitation was explicitly adjusted during online ML-corrected runs: the column-integrated ML humidity tendency was added to the surface precipitation budget (their Eq.~2) to prevent spurious global drying of the land surface. In our implementation, no post-processing correction was applied; yet, improvements in precipitation were still achieved. As discussed in~\nameref{sec:offline_results}, one possible explanation is the advantage of deep-learning architectures employed here, compared to the random forest model used by Watt-Meyer et al. (2021)$^{\cite{watt2021correcting}}$. Finally, we note that UNetMP, despite substantially increasing the parameter count relative to the baseline UNet, exhibited systematic degradation, particularly in upper-tropospheric winds, temperature, and humidity (e.g., at 200~hPa). These degradations were accompanied by declines in cloud water properties (e.g., LWP and IWP) and several radiative fluxes, indicating that additional parameters do not necessarily translate into improved performance.

\begin{figure}[ht]
    \centering
    \includegraphics[width=\linewidth]{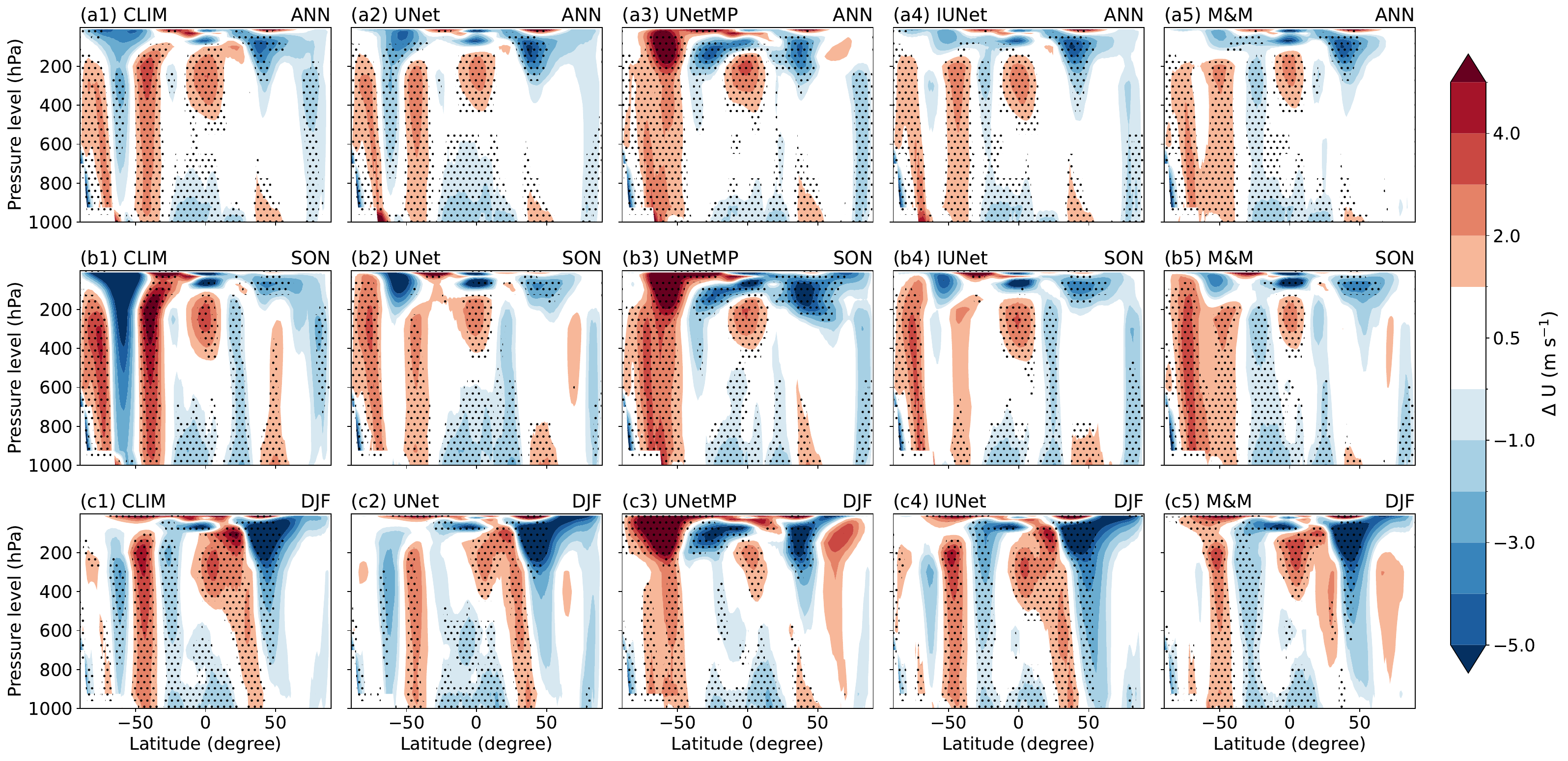}
    \caption{\textbf{Zonally averaged biases in zonal wind ($\Delta U$; m\,s$^{-1}$) over 2012–2016}. Shown are the annual mean (ANN; panels a1--a5), boreal autumn (SON; panels b1--b5), and boreal winter (DJF; panels c1--c5). Biases are computed relative to ERA5 reanalysis for the same period. Columns correspond to different model configurations: free-running EAMv2 (CLIM, first column) and EAMv2 with online ML bias correction using UNet (second), UNetMP (third), IUNet (fourth), and M\&M (fifth). Shading indicates the mean bias (EAMv2 minus ERA5). Stippling marks regions where the bias is statistically significant at the 95\% level, based on a two-sided $t$-test with effective sample size adjusted for temporal autocorrelation using a first-order autoregressive [AR(1)] model$^{\citep{Santer2000}}$. Pressure (hPa) is shown on the vertical axis and latitude (degrees) on the horizontal axis. Details of the simulation configurations are provided in Supplementary Table~S5.}
    \label{fig:zonal_bias_U}
\end{figure}

To gain further insight into the behavior of the ML bias correction models, Figure~\ref{fig:zonal_bias_U} shows the zonally averaged biases in the zonal wind field ($\Delta U$) for the annual mean (ANN; panels a1--a5) as well as for the two seasons that exhibited the greatest (SON; panels b1--b5) and least (DJF; panels c1--c5) RMSE reductions in Figure~\ref{fig:rmse_heatmap_5yr}. The effectiveness of the ML bias correction models clearly depends on both latitude and season. The proposed ML architectures, except for UNetMP, consistently reduced biases throughout most of the mid- and lower troposphere, with the strongest improvements during SON, driven largely by reductions in the Southern Hemisphere storm-track region (45$^\circ$--65$^\circ$S). In contrast, all models showed a limited ability to reduce biases in the upper troposphere (above 200~hPa), particularly during DJF, as substantial errors in the CLIM simulation (panel c1) were largely unchanged after correction for ML bias (panels c2--c5). When comparing across the four ML architectures, UNetMP (panels a3--c3), despite having substantially more parameters than the baseline UNet (panels a2--c2), showed systematic degradation, primarily through overcorrections of biases in the SH storm-track region (45$^\circ$--65$^\circ$S). Because these regions are characterized by intense large-scale variability driven by synoptic eddy activity, and further modulated by mesoscale air–sea coupling and diabatic processes$^{\citep{hoskins1990stormtracks,minobe2008sst,pfahl2015moist}}$, this outcome is consistent with the intended purpose of increasing the number of trainable parameters in UNetMP: to enhance representational capacity and capture more complex correction patterns arising from these processes. However, rather than improving performance, the additional parameters appear to have amplified model biases, highlighting that increased complexity does not necessarily translate into more effective corrections. By contrast, IUNet and M\&M, which had comparable parameter counts but employed different architectural designs, did not exhibit such degradations. This indicates that it is not the parameter count alone but rather how parameters are structured and utilized within the network that determines performance. A similar conclusion regarding the performance of the ML bias correction models can be drawn from the zonally averaged temperature biases, where improvements are more spatially uniform across latitudes and M\&M provides the most robust corrections (see Supplementary Figure~S11).

\begin{figure}[ht]
\centering
    \includegraphics[width=\linewidth]{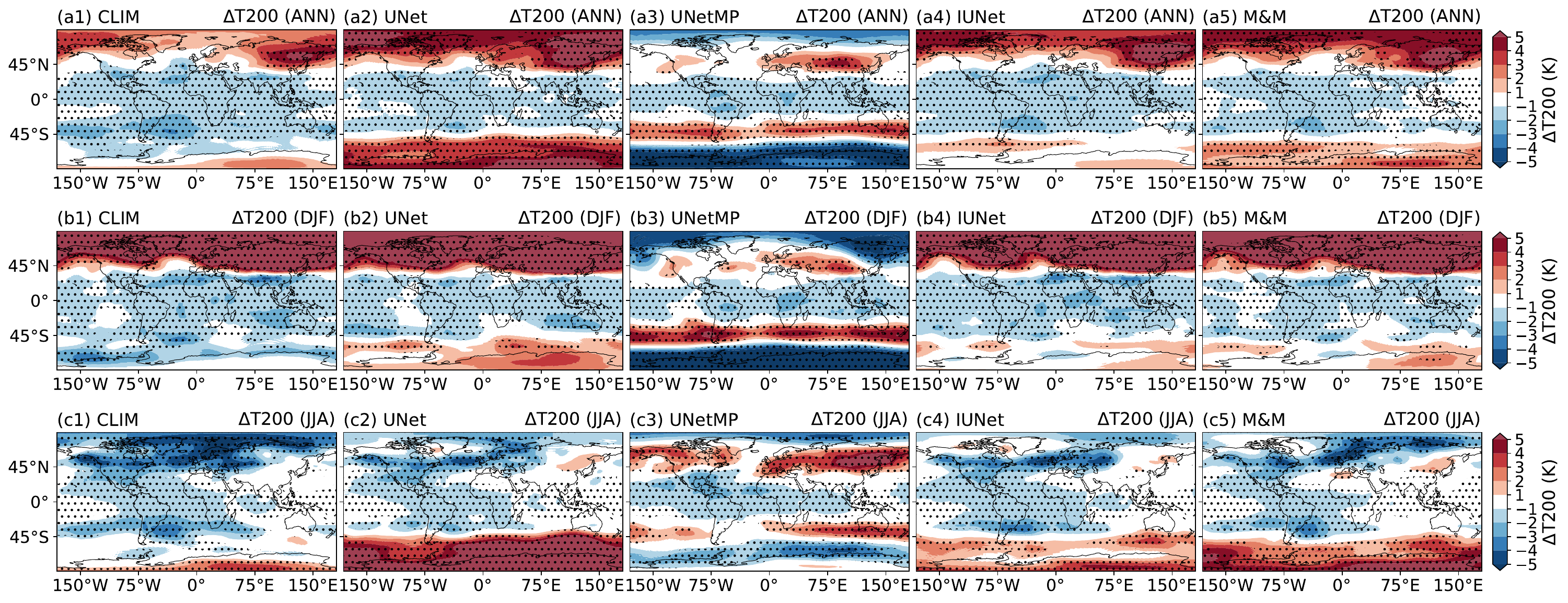}
    \caption{\textbf{Global 200-hPa temperature biases ($\Delta T_{200}$; K) relative to ERA5, averaged over 2012--2016.} Rows correspond to (a) annual mean (ANN), (b) boreal winter (DJF), and (c) boreal summer (JJA). Columns show the baseline simulation (CLIM) and online ML bias-corrected experiments (UNet, UNetMP, IUNet, and M\&M) with EAMv2. Shading indicates mean bias (simulation minus ERA5) using a diverging colormap (red = warm, blue = cold). Stippling marks regions where the bias is statistically significant at the 95\% confidence level, based on a two-sided $t$-test with effective sample size adjusted for temporal autocorrelation using a first-order autoregressive [AR(1)] model$^{\citep{Santer2000}}$. Color bars denote the bias range in Kelvin. Details of the simulation configurations are provided in Supplementary Table~S5.}
\label{fig:temp_bias_map}
\end{figure}

Based on the above discussion, Figure~\ref{fig:temp_bias_map} further presents the global distribution of 200-hPa temperature biases for the annual mean (ANN) as well as for boreal winter (DJF) and summer (JJA). In the ANN mean (Figure~\ref{fig:temp_bias_map}a1), the free-running EAMv2 simulation (CLIM) exhibits warm biases over the Arctic and Antarctic continental regions, contrasted with widespread cold biases elsewhere, with the strongest negative anomalies centered over the Southern Ocean. At NH high latitudes, especially the Arctic, biases show a pronounced seasonal contrast---warm during DJF (Figure~\ref{fig:temp_bias_map}b1) and cold during JJA (Figure~\ref{fig:temp_bias_map}c1). These opposing signals partially offset one another in the annual mean, yielding smaller but still substantial residual biases. In contrast, SH high-latitude biases are relatively weaker and show little seasonal compensation, with persistent warm anomalies over Antarctica accompanied by strong cold anomalies over the Southern Ocean (Figs.~\ref{fig:temp_bias_map}a1--c1). For the ML bias-corrected simulations, all four architectures show a weaker but systematic reduction of cold biases across the tropics and subtropics relative to CLIM, but their performance diverges between hemispheres.

In the NH annual mean, UNet, IUNet, and M\&M introduce degradations, producing amplified warm biases relative to CLIM (Figs.~\ref{fig:temp_bias_map}a1--a2 and a4--a5). This degradation arises in part because the seasonal error compensation present in CLIM is no longer active: while these models effectively reduce the JJA cold temperature biases (Figs.~\ref{fig:temp_bias_map}c1--c2 and c4--c5), they fail to correct the pronounced DJF warm temperature biases (Figs.~\ref{fig:temp_bias_map}b1--b2 and b4--b5), resulting in a more pronounced residual warm bias in the ANN mean. In the SH annual mean, UNet exhibits a significant degradation, producing amplified warm biases relative to CLIM (Figs.~\ref{fig:temp_bias_map}a1--a2). This primarily reflects its poor performance in JJA, where the model introduces substantial warm biases over both the Southern Ocean and the Antarctic continent (Figs.~\ref{fig:temp_bias_map}c1--c2). In contrast, IUNet and M\&M, owing to their different architectures, substantially mitigate the unintended biases seen in UNet (Figs.~\ref{fig:temp_bias_map}a1--c5), although they still exhibit weak overcorrection of JJA biases compared to those in CLIM (Figs.~\ref{fig:temp_bias_map}c1--c5). Relative to the other three architectures, UNetMP again exhibits distinct behavior. Its relatively smaller ANN biases represent an apparent improvement that is, however, largely attributable to compensating errors between seasons, most notably between DJF and JJA (Figs.~\ref{fig:temp_bias_map}a3--c3). When comparing UNet and UNetMP with CLIM, the ML bias correction appears to under-correct with $\sim$0.2M parameters (UNet; Figs.~\ref{fig:temp_bias_map}a2--c2), while it tends to overcorrect with $\sim$6.7M parameters (UNetMP), as indicated in part by the flipped sign of high-latitude errors during both DJF and JJA. These results collectively underscore the need for caution when increasing parameter size without adapting the underlying architecture, since excessive representational capacity and overly intricate correction patterns can ultimately degrade the effectiveness of ML bias correction. Moreover, the four ML bias correction models, distinguished by parameter size and architectural design, fail to deliver uniform improvements across all regions, vertical layers, and variables. Their performance is highly sensitive to latitude, season, and variable type, with pronounced degradations in the upper troposphere and high latitudes, even as biases are clearly reduced in the tropics and subtropics. These results point to the potential benefits of combining architectures tailored to specific regions or variables, a strategy that could yield more balanced corrections but remains an open question for future investigation. Finally, all ML bias-correction models showed notable difficulties in the polar regions during sea-ice formation seasons, with degradation most pronounced in the upper troposphere (Figs.~\ref{fig:temp_bias_map}b1--b5 and c1--c5) but absent near the surface (see Supplementary Figure~S12). While sea-ice concentration was incorporated into the conditioning scheme (see~\nameref{sec:scalar_embed}), this input alone was insufficient to capture the complex upper-tropospheric biases. Future work exploring dynamic conditioning strategies that represent the evolving sea-ice state may help address these limitations.

Three factors help explain the imperfect performance of our proposed ML bias correction models in the upper troposphere, particularly over high-latitude regions. First, the substantial background biases in E3SMv2 over the NH high latitudes constrain the ability of the ML models to achieve effective corrections. In this region, biases exhibit a pronounced seasonal contrast, with warm anomalies during DJF and cold anomalies during JJA. This strong seasonal sign reversal reflects the intrinsic temporal variability of the model errors, which poses a significant challenge for the ML models to learn stable and effective correction mappings. Second, as noted in our previous work on offline ML bias correction$^{\cite{barthel2024debiasing}}$, atmospheric variables (e.g., wind and temperature) in the upper troposphere are characterized by intrinsically low signal-to-noise ratios and greater observational uncertainty, which limit the ability of ML models to extract robust correction patterns, particularly when training datasets are relatively short. Consistent with this, the models presented in this study---trained on only two years (2009--2010)---were especially prone to such challenges. Third, variability associated with large-scale atmospheric modes---such as the El Ni\~{n}o--Southern Oscillation (ENSO)---can strongly influence upper-tropospheric circulation and its vertical coupling with thermodynamic fields. Insufficient representation of this variability in the training data reduces the generalization capability of ML models. For example, a record-breaking ENSO event occurred during 2015--2016$^{\citep{LHeureux2017}}$ within our test period. As shown in Supplementary Figure~S10, which presents the same RMSE metrics as Figure~\ref{fig:rmse_heatmap_5yr} but excludes 2015--2016, the RMSE reductions across most quantities are much clearer. Because our training period (2009--2010) did not include such strong ENSO variability, the 2015--2016 event effectively represents an out-of-sample case for our ML bias correction models, where degraded performance would be expected. Taken together, these factors underscore the need to extend training datasets to encompass a broader range of atmospheric variability and to refine ML model design to enhance robustness and generalization.

Overall, our results in this section demonstrate that ML-based bias correction can be stably and effectively implemented in an online EAMv2 framework. All models capture the principal spatial structures of correction tendencies and achieve notable reductions in RMSE (2--10\%) across most variables and seasons. IUNet and M\&M provide the most consistent gains, while the comparison between UNet and UNetMP underscores that parameter growth without architectural refinement may hinder, rather than improve, performance. Challenges remain in correcting upper-tropospheric winds and polar-region biases, reflecting both intrinsic variability and limited training data. Nevertheless, the consistency with offline results and the improvement of precipitation without post-processing highlight the robustness of our framework, establishing a strong foundation for future refinement through extended training datasets and enhanced conditioning strategies.

\section{METHOD} \label{sec:method}
\subsection{\textbf{Problem Setup}}
Earth system models such as E3SM \cite{doecode_10475} exhibit systematic atmospheric biases arising from limited resolution, approximations in model formulation, and parameterizations. Recent hybrid ML–nudging strategies \cite{watt2021correcting,Bretherton2022} address this by learning from controlled, nudged simulations: an ML model is trained to reproduce the nudging tendencies that steer the model state toward a reference, and the learned correction is then applied during free-running simulations without real-time observational constraints. 

In this study, we adopt the same principle for E3SM’s atmosphere: given an instantaneous model state \(X_m(t)\) (e.g., winds, temperature, humidity) and the corresponding reference-consistent nudging tendency \(\big(\partial X_m/\partial t\big)_{\mathrm{ndg}}(t)\), we learn a compact neural operator \(f_\theta\) that predicts a single-step bias-correction tendency,
\begin{equation}
    \Big(\tfrac{\partial X_m}{\partial t}\Big)_{\!\mathrm{ndg}}(t) = f_\theta\!\big(X_m(t),\,\mathcal{M}(t)\big),
\qquad
\bar X_m(t) = X_m(t) + \Big(\tfrac{\partial X_m}{\partial t}\Big)_{\!\mathrm{ndg}}(t),
\end{equation} so that the corrected state \(\bar X_m(t)\) converges toward the reference (Eq.~\eqref{eq3}). The metadata \(\mathcal{M}(t)\) encodes static or slowly varying geophysical context (e.g., orography, land–sea mask, latitude) supplied via FILM embeddings. 

Two practical constraints motivate this \emph{single-step, history-free} formulation with a compact architecture: during online coupling only the immediately previous timestep is exposed at inference, and production runs are CPU-bound. Training uses paired samples \(\big(X_m(t), (\partial X_m/\partial t)_{\mathrm{ndg}}(t)\big)\) with per-channel normalization to handle heterogeneous variable scales; evaluation employs complementary spatial and temporal diagnostics (pattern and time correlations, error and structure metrics such as MSE/RMSE/MAE, PSNR, bias, SSIM, and global \(R^2\)). Detailed data splits, architectural variants, and coupling protocols are described in subsequent subsections.
\subsection{\textbf{Training Data Construction}} \phantomsection
\label{sec:training_data} 
The training dataset for our ML model is derived from EAM simulations that employ a nudging framework, as described in previous studies \cite{Zhang_et_al:2022}. In brief, nudging introduces a corrective tendency term into the model equations at each time step to guide the evolution of a physical variable $X_m$ toward a specified reference state. The governing equation with nudging can be written as:
\begin{equation} \label{eq1}
    \frac{\partial X_m}{\partial t} = D(X_m) + F(X_m) + \left(\frac{\partial X_m}{\partial t}\right)_{\mathrm{ndg}},
\end{equation}
where $X_m$ denotes the model state variable, and $D(X_m)$ and $F(X_m)$ represent the contributions from the model’s dynamical core and physical parameterizations, respectively. The final term is the nudging tendency, which, in discrete form, is given by:
\begin{equation} \label{eq2}
    \left(\frac{\partial X_m}{\partial t}\right)_{\mathrm{ndg}} = - \frac{X_p(t) - X_m(t)}{\tau},
\end{equation}
where $\tau$ is the nudging timescale that controls the strength of the relaxation, which is set to 6 hours following Zhang et. al. (2022) \cite{Zhang_et_al:2022}. $X_p(t)$ is the reference (or prescribed) value of the state variable at time $t$.

For the nudged training dataset used in this study, selected model state variables—specifically horizontal winds (U, V), temperature (T), and specific humidity (Q)—are continuously relaxed toward reference values from the fifth-generation reanalysis produced by the European Centre for Medium-Range Weather Forecasts (ERA5). The ERA5 reanalysis data were collected at 3-hourly frequency and remapped to the EAM model grid. Nudging tendencies were computed at every 30-minute model time step following Eq.~\eqref{eq2}, where \( X_p \) is obtained by linearly interpolating the 3-hourly ERA5 data to the model time.

We note that nudging of specific humidity has been reported to significantly disrupt the simulated hydrological cycle, introducing artificial moisture sources and sinks. This, in turn, can degrade the physical realism of precipitation and moisture processes, thereby contaminating the training signal for ML models. To address this issue, we adopted a revised nudging strategy as described in Zhang et al.\ (2025)~\cite{Zhang_et_al:2025}, which has been shown to better preserve the integrity of the hydrological cycle and maintain realistic precipitation patterns in EAM. A detailed discussion of the nudging strategy design is beyond the scope of this paper; interested readers are referred to Zhang et al.\ (2025)~\cite{Zhang_et_al:2025} for further information.

Table~\ref{tab:ml_variables} summarizes the variables used for ML training, which are further discussed in the next subsection. The model state variables were archived as instantaneous outputs at the beginning of each 3-hourly window. In contrast, the corresponding nudging tendencies were computed as 3-hourly averages to suppress high-frequency noise and better capture the effective tendencies applied during model integration. Following the methodology of Zhang et al.\ (2025)~\cite{Zhang_et_al:2025}, nudging tendencies in certain layers, particularly within the planetary boundary layer and near the top of the model, were excluded due to concerns about the reliability of the reference data and the potential interference with the model’s physical parameterizations. As will be discussed in the next section, these layers were also excluded from ML  training and prediction.
\begin{table}[htbp]
\centering
\scriptsize
\caption{List of variables used for ML training.}

\begin{tabularx}{\textwidth}{lXlll}
\hline
\textbf{Variable} & \textbf{Description} & \textbf{Dimension} & \textbf{Excluded Levels} & \textbf{Frequency} \\
\hline
U         & Zonal wind component (m/s)               & (lev, lat, lon)   &  1-10        & 3-hourly  (instantaneous) \\
V         & Meridional wind component (m/s)          & (lev, lat, lon)   &  1-10                          & 3-hourly  (instantaneous) \\
T         & Air temperature (K)                      & (lev, lat, lon)   & 2                          & 3-hourly  (instantaneous) \\
Q         & Specific humidity (kg/kg)                & (lev, lat, lon)   & 2                        & 3-hourly  (instantaneous) \\
\hline
UTEND     & Nudging tendency for U (m/s$^2$)         & (lev, lat, lon)   &  1-10                          & 3-hour mean \\
VTEND     & Nudging tendency for V (m/s$^2$)         & (lev, lat, lon)   &  1-10                           & 3-hour mean \\
TTEND     & Nudging tendency for T (K/s)             & (lev, lat, lon)   &  1-10                           & 3-hour mean \\
QTEND     & Nudging tendency for Q (kg/kg/s)         & (lev, lat, lon)   & 1-10                        & 3-hour mean \\
\hline
LANDFRAC  & Land fraction (0--1)                     & (lat, lon)        & --                        & Static \\
ICEFRAC   & Sea ice fraction (0--1)                  & (lat, lon)        & --                        & Static \\
OCNFRAC   & Ocean fraction (0--1)                    & (lat, lon)        & --                        & Static \\
Lat       & Latitude (degrees north)                 & (lat)             & --                        & Static \\
Lon      & Longitude (degrees east)                 & (lon)             & --                        & Static \\
lev       & Model vertical level index               & (lev)             & --                        & Static \\
area      & Grid cell area (m$^2$)                   & (lat, lon)        & --                        & Static \\
COSZ      & Cosine of solar zenith angle             & (lat, lon)        & --                        & 3-hourly (Instantaneous)  \\
PHIS      & Surface geopotential (m$^2$/s$^2$)       & (lat, lon)        & --                        & Static\\
\hline
\end{tabularx}
\label{tab:ml_variables}
\end{table}

The nudged simulations of EAMv2 used to generate the machine learning training data in Table~\ref{tab:ml_variables} included the active atmosphere and the land components of E3SM, while prescribing the sea surface temperature (SST) and the sea ice concentration (SIC), consistent with the protocol of the Atmospheric Model Intercomparison Project (AMIP) \cite{Gates_et_al:1999}. The SST and SIC data are monthly values sourced from the input4MIPs datasets \citep{Reynolds_2002_SST}, ensuring alignment with the AMIP framework. Other external forcing data, including volcanic aerosols, solar variability, greenhouse gas concentrations, and anthropogenic emissions of aerosols and their precursors, were prescribed according to the protocols of the World Climate Research Programme (WCRP) Coupled Model Intercomparison Project Phase 6 (CMIP6) \cite{Eyring_et_al:2016, Hoesly_et_al:2018, Feng_et_al:2020}. Emissions of aerosols and their precursor gases from the year 2010 were used to represent present-day conditions.

% ===================== METHODS =====================
\subsection{Scalar embedding via FiLM conditioning} \label{sec:scalar_embed}
We encode slowly varying geophysical context via FiLM modulation. The conditioning vector includes LANDFRAC, ICEFRAC, OCNFRAC, Latitude, Longitude, lev, area, COSZ, PHIS, most of which are time-invariant. A shared embedding network (MLP+CNN) maps this metadata $\mu$ to per-level affine parameters $(\gamma_k,\beta_k)$. Given intermediate features $F_k$ from the base model $M(X_m)$, modulation is applied as
\begin{equation}
    \mathrm{FiLM}(F_k;\gamma_k,\beta_k) \;=\; \gamma_k \odot F_k \;+\; \beta_k,
\end{equation} with $(\gamma_k,\beta_k)$ broadcast to the corresponding stage.
\noindent\textit{Level-wise application across all UNet variants.}
As illustrated in Figure~\ref{fig:1}D (bottom), each resolution level (Block~0\,--\,Block~4) of all UNet-based architectures UNet, IUNet, and M\&M receives its own FiLM parameters produced by a FilmGenerator1D. Where branches are context-dependent, the generator yields $(\gamma'_k,\beta'_k)$ for the appropriate path. This modular, learnable conditioning at every encoder/decoder level enables dynamic, depth-aware adaptation while preserving invertible affine structure (diagonal Jacobian $1+\gamma_k>0$), and integrates static spatial context without requiring temporal history at runtime.

\subsection{Neural operator architecture} \label{sec:ml_methodology}
The neural operator learns the mapping from the model state to the nudging correction. Let us say $X_{\text{m}}$ is the state of the model and $\left(\frac{\partial X_m}{\partial t}\right)_{\mathrm{ndg}}$ be the nudging correction for that particular state, such that\begin{equation}\bar{X}_{\text{m}} = X_{\text{m}} + \left(\frac{\partial X_m}{\partial t}\right)_{\mathrm{ndg}} \label{eq3},
\end{equation} 
where $\bar{X}_{\text{m}}$ denotes the corrected model state which should be close to ERA5. By construction, \eqref{eq3} tries to converge the prediction from the model towards the true observations (here ERA5). The training strategy for online bias correction faces two constraints: (i) deployment within E3SM exposes only the immediately previous timestep, so the model cannot maintain runtime history or exploit longer past context; and (ii) E3SM runs on CPUs, ruling out large, high-parameter models. Moreover, enabling access to longer temporal histories would require substantial changes to E3SM’s framework and would likely introduce compute bottlenecks. Given that predictions must be made using only a single time step, framing the task as an image-to-image prediction is a natural choice. Consequently, we adopt a UNet architecture as the foundation for our neural operator framework. We also adopt per-channel normalization to address heterogeneous feature scales, which empirically improves optimization and accelerates convergence.  Traditional UNets excel at dense sequence prediction by combining a contracting path of repeated “double‐conv + downsample” stages with a symmetric expanding path and skip-connections, thereby blending local detail and global context. Building on this, we developed IUNet, a new architecture that replaces each fixed double-conv with an InceptionBlock1D (parallel 1×1, 3×1, 5×1, and pooled branches fused to capture multiple spatial scales) and injects learned metadata embeddings at every layer, giving each block explicit access to geophysical context alongside the primary signal. M \& M, our next advance, then addresses further limitations of upsampling. So instead of a single interpolation or transpose-conv, it employs a three-branch UpMultiBlock1d (transpose-convolution, inception-style interpolation, and pixel-shuffle) whose concatenated linear maps span the entire low-resolution feature space. We show in  Lemma 2.3.1 how this multi-branch design yields a full-rank, injective upsampling operator—preserving every input direction—and a subsequent 1×1 fusion retains that injectivity, ensuring uniquely invertible, artifact-free reconstructions at each decoder stage. \subsubsection*{UNet} \label{sec:UNet}
We adopt the UNet architecture originally proposed by Ronneberger et al. [1] as our core feature extractor and reconstructor. UNet is an encoder–decoder network with symmetric skip-connections that has proven effective for dense prediction tasks, such as image segmentation and regression, by jointly capturing context and precise localization. The encoder consists of a sequence of four downsampling blocks. Each block comprises two consecutive 3×3 convolutional layers (stride = 1, padding = 1), each followed by a rectified linear unit (ReLU) activation. After the second convolution, a 2×2 max-pooling operation with a stride = 2 reduces the spatial resolution by half, while the number of feature channels is doubled. This path extracts increasingly abstract representations, enabling the network to learn rich semantic features at multiple scales. The decoder mirrors the encoder’s structure in reverse. Each upsampling block begins with a 2×2 transposed convolution that doubles the spatial dimensions and halves the number of channels. The up-sampled feature map is concatenated with the correspondingly sized feature map from the encoder via a skip-connection, restoring fine-grained spatial details. Two further 3×3 convolutions (each followed by ReLU) then refine the combined features. A 1×1 convolution maps the last channel feature map to the desired number of output channels. Crucially, the UNet’s skip-connections shuttle high-resolution features from the encoder directly into the decoder. This design mitigates information loss during pooling and allows the network to precisely localize structures even while leveraging deep, context-rich representations. In our experiments, we initialize all convolutional kernels with the normal initialization, and insert batch-normalization layers after each convolution to accelerate convergence and improve stability. We apply a dropout of 0.2 in the bottleneck to regularize against overfitting. The network is trained end-to-end using the Adam optimizer with an initial learning rate of $5\times 10^{-4}$ and weight decay of $1\times 10^{-5}$. Figure \ref{fig:1} (C)-(i) (a) shows a schematic figure for the UNet that we use for our problem setup. 

\subsubsection{IUNet} \label{sec:IUNet}
IUNet builds on the classic UNet by replacing each “double‐conv” stage with a four‐branch InceptionBlock1D \cite{szegedy2016rethinking}.  In practice, each Inception block first projects the input via a 1x1 convolution, then simultaneously applies three kernels of size 3 and 5 (each prefaced by its own 1x1 bottleneck) plus a pooled‐feature pathway; these four responses are concatenated, normalized, and activated, giving the network immediate access to fine, medium, and coarse spatial patterns as well as invariant summaries. After each Inception stage, the metadata—linearly interpolated to match the current feature length—is embedded by two 1x1 convolutions and concatenated onto the feature map, ensuring that every layer can condition its filters on geophysical context. The encoder comprises four such blocks interleaved with max‐pooling to halve resolution (and double channels from 64 up to 512), followed by a two‐block bottleneck at 1024 channels. The decoder mirrors this structure with transpose‐convolutions for upsampling, skip connections from the encoder, repeated Inception–metadata injections, and a final 1X1 convolution to produce the desired output channels, which are then reshaped and denormalized back to physical units. Compared to a vanilla UNet’s fixed‐kernel convs, IUNet’s multi-scale Inception blocks capture richer spatial hierarchies, metadata injection provides explicit, continuous context, and FiLM offers more flexible, reversible conditioning—all combining to deliver sharper, more physically consistent 1D sequence predictions. Figure \ref{fig:1} (C)-(ii) shows a schematic figure for the IUNet that we use for our problem setup.

\subsubsection{M\&M}  \label{sec:MnM}
The M\&M network begins by normalizing each $x$-channel to the range $[-1, 1]$, concatenating the sliced secondary signal $y$, and reshaping the result into a 1D feature tensor. The encoder then applies a subsequent block of DoubleConv1d blocks (two $3 \times 1$ convolutions followed by BatchNorm and ReLU activations) and DownMultiBlock1d modules—each fusing strided convolution, max-pooling + convolution, and average-pooling + convolution paths—to successively halve the resolution while doubling the channel depth. At the bottleneck, a final DoubleConv1d expands the feature richness, after which the decoder mirrors this process. Rather than using a single upsampling operator as in a standard UNet, which corresponds to one linear map of limited rank and may collapse certain low-resolution directions at each decoder stage, M\&M uses our three-branch UpMultiBlock1d:
\begin{itemize}
    \item ConvTranspose1d: This branch uses a learned transpose‐convolution to simultaneously insert new samples and convolve them with trainable kernels. 
    \item InceptUpsample1d: (linear interpolation + parallel $3$, $5$, and $7$-kernel convolutions) Here, we first apply a fixed linear interpolation to guarantee a smooth baseline doubling of the sequence length. We then pass the interpolated signal through three parallel convolutions of sizes 3, 5, and 7, each capturing features at increasingly broader contexts. A final 1x1 projection adaptively fuses these multi-receptive-field responses.
    \item PixelShuffle1d: This branch begins by expanding the channel dimension by the upsampling factor using a 1x1 convolution, then reorganizes (or “shuffles”) those extra channels into the spatial dimension to double the length. Because it avoids any learned spatial kernel, it inherently sidesteps checkerboard artifacts and decouples channel mixing from sample insertion.
\end{itemize}

By using Lemma 2.3.1. (see Supplementary Section~S1),stacking these three complementary operators yields a block whose rank exceeds that of any single branch. When their row-spaces are chosen appropriately, the combined block reaches full column rank. In practice, this means our upsampling strategy has a lower chance of losing any low-resolution features. A subsequent $1 \times 1$ fusion layer with sufficient output channels preserves this injectivity, ensuring high-fidelity reconstruction at all scales. Figure \ref{fig:1} (C)-(iii) shows a schematic figure for the M\&M that we use for our problem setup.

\section{CONCLUSION}
This study explored neural network architectures for bias correction in the E3SM atmosphere model (EAMv2). We constructed mapping functions between instantaneous model states of wind ($U,V$), temperature ($T$), and humidity ($Q$), and their corresponding 3-hour mean nudging tendencies, using three FiLM-conditioned deep learning methods: UNet, IUNet, and M\&M. The performance of the ML bias correction models were evaluated through both offline tests and online integrations.  \newline

Trained on 2009-2010 and evaluated on 2015, the models were assessed with complementary diagnostics (bulk temporal correlation, layer-wise time correlation, and mean pattern correlation) and standard error/structure metrics (MSE, RMSE, MAE, PSNR, Bias, StdError, global \(R^{2}\), CV error, SSIM), all targeted to reproducing the \emph{mean} nudging tendency that accumulates in the online E3SM loop. At a matched parameter budget (\(\sim6.7\)M), \textbf{M\&M} delivers the best overall skill for winds \(U\) and \(V\): lowest errors (MSE/RMSE/MAE), highest PSNR/SSIM, strongest \(R^{2}\), and near-zero bias, together with the most uniform, positive layer-wise time correlations across the column. For humidity \(Q\), \textbf{IUNet} is top on most metrics and spatial agreement (consistent with its inception multi-path advantage on \(Q\)'s finer, fragmented structure), while \textbf{M\&M} is a close second with excellent temporal fidelity (high, stable layer-wise correlations without negatives). For temperature \(T\), \textbf{M\&M} attains the lowest errors and highest PSNR and \(R^{2}\), and the most consistent vertical temporal coherence; the small UNet’s higher mean pattern correlation is attributable to a low-pass smoothing bias (beneficial for pattern correlation but not for absolute error). Training times per epoch remain lightweight (milliseconds), with \textbf{M\&M} only modestly slower than IUNet/UNet, so accuracy gains come with minimal overhead. Because corrections are injected at every model step (or every six steps), even small offline improvements compound. Collectively, these offline results demonstrate robust out-of-year generalization in both spatial structure and vertical phase tracking, providing strong evidence that the proposed approach should reduce long-horizon climate bias when coupled online.

The online experiments over 2012–2016 demonstrate that ML-based bias correction can be stably integrated in a 5-year long-term EAMv2 simulations, with measurable benefits relative to the free-running baseline simulation. All models reproduced the large-scale structures of nudging tendencies and reduced global mean biases across multiple variables and seasons, including zonal-mean winds, temperature, humidity, and surface fluxes. Among the tested architectures, IUNet and \textbf{M\&M} consistently achieved the largest improvements, yielding 2–10\% reductions in RMSE for key thermodynamic and dynamical fields and more robust corrections across regions and seasons. By contrast, the parameter-increased UNetMP did not outperform the smaller UNet, underscoring that increased complexity alone does not guarantee better online performance. Overall, the results highlight that carefully designed architectures with inductive biases (e.g., IUNet’s multi-path design, \textbf{M\&M}’s operator learning) provide the most stable and effective corrections, offering a foundation for hybrid approaches that combine physical modeling with data-driven methods to improve the fidelity of Earth system simulations.

At the same time, several limitations emerged. Although RMSE reductions were evident, the improvements were generally modest, confined to the mid–lower troposphere, and spatially heterogeneous. Substantial biases persisted in the upper troposphere and polar regions, where the models struggled with seasonal sign reversals in the NH and entrenched systematic biases in the SH. These deficiencies likely result from three main factors.  First, as discussed in~\nameref{sec:online_perofrmance}, deficiencies in our training setup, particularly the relatively short training period, may have limited the ability of the ML models shown in this study to learn the wide-ranging corrections that vary across seasons, latitudes, and anomalous years (e.g., ENSO). Second, ML models were trained offline on nudging-derived data and, when deployed online, are exposed to distributional shifts, a lack of feedback during training, and potential violations of physical constraints—limitations that have also been reported in previous studies \citep{brenowitz2018prognostic,rasp2018deep,chantry2021machine}. Third, unlike nudging, which applies continuous corrections based on deviations from a reference state, ML corrections rely solely on the internally predicted state of the model and are applied as terms with intermittent weak tendency to preserve numerical stability. This design inherently limits corrective strength, particularly for persistent or dynamically coupled biases such as those in the upper troposphere. We are currently exploring the extended training of the ML architectures proposed in this study, using longer and more diverse datasets to capture variability more effectively. While promising, these efforts are ongoing and are not reported here. In contrast, the remaining two challenges will require more substantial advances, including the development of online learning strategies that better assimilate ML-predicted tendencies and conditioning schemes that more explicitly encode boundary processes such as sea-ice dynamics. Nevertheless, the framework presented in this study provides a foundation for these next steps, where more robust online learning strategies, extended training datasets, and improved conditioning schemes can be systematically tested and refined.

Finally, high-resolution Earth system models are indispensable for advancing the physical and dynamical realism of simulated processes that govern high-impact weather and climate variability \citep{Haarsma2016, Bryan2019, Roberts2020}. Yet, the substantial computational demands associated with increased fidelity remain a major barrier to broader application \citep{Balaji2017}. For instance, refining horizontal resolution from approximately 100~km to 25~km in E3SM entails nearly a 64-fold increase in computational cost owing to the cubic scaling of spatial and temporal resolution \citep{Held2019, Bacmeister2014, Caldwell2019}. Results of online model throughput (See Supplementary Figure~S13) illustrate that the proposed architectures maintain feasible computational performance, with simulated years per day (SYPD) ranging from 1.3 for the more complex architecture (IUNet) to 8.5 for UNet, compared with 12.7~SYPD for the baseline CLIM configuration—corresponding to only about 1.5–10 times slower performance relative to CLIM, in stark contrast to the 64-fold cost increase required for a fourfold resolution refinement. The offline inference cost of the ML models is only a fraction of a second on GPU, but the current E3SM only runs on CPUs. Pairing online bias correction with a GPU-enabled earth system model would largely reduce the performance penalties seen here. In this context, these ML-based frameworks offer an emerging alternative pathway to enhance model fidelity at greatly reduced computational expense. Although the present implementation focuses on demonstrating online stability and computational feasibility rather than delivering substantial bias reduction, the approaches introduced here establish a proof of concept for robust and stable ML integration within a coupled Earth system model.

\section{DATA AVAILABILITY}
The data that has been used in the findings of this research work are available from the corresponding authors upon reasonable request.

\section{CODE AVAILABILITY}
The code that has been used in the findings of this research work is available from the corresponding authors upon reasonable request.

\section{Acknowledgements}
%%%%%%%%%%%%%%%%%%%%%%%%%%%%%%%%%%%%%%%%%%%%%%%%%%%%
The authors acknowledge Dmitry Alexeev for his help with the PyTorch Fortran interface for integrating the ML models with E3SM. The authors would also like to acknowledge the use of the CCV's Oscar HPC cluster at Brown University. The authors also acknowledge the use of ChatGPT (OpenAI) to refine the wording of the manuscript.

\noindent \textbf{Funding:} 
This work was supported by the Scalable, Efficient and Accelerated Causal Reasoning Operators, Graphs and Spikes for Earth and Embedded Systems (SEA-CROGS) project, (DE-SC0023191), funded by the U.S. Department of Energy Advanced Scientific Computing Research (ASCR) program. PNNL is operated for the Department of Energy by Battelle Memorial Institute under contract DE‐AC05‐76RL01830.

\section*{Author contributions statement}
A.B. and S.Z. : 
Conceptualization, Methodology, Software, Formal Analysis, Investigation, Data Curation, Writing - Original Draft, Writing - Review \& Editing and Visualization.
K.S.:
Methodology, Software, Formal Analysis, Investigation, Writing - Original Draft and Writing - Review \& Editing
B.H.:
Investigation, Validation, Resources, Writing - Original Draft and Writing - Review \& Editing
G.E.K. and L.R.L.:
Conceptualization, Validation, Formal Analysis, Resources, Writing - Review \& Editing, Supervision, Project administration and Funding acquisition.
\section*{Additional information}
% \textbf{Accession codes} The codes and data developed in this study are made available to the Editor to be passed on to the reviewers \href{https://github.com/vivekoommen/Gen4Turbulence}{GitHub Repository}\cite{Gen4Turbulence2025}.

\noindent\textbf{Competing interests} The authors declare no competing interests.

\bibliography{reference} % No styEntries are in the refs.bib file

\begin{thebibliography}{10}
\urlstyle{rm}
\expandafter\ifx\csname url\endcsname\relax
  \def\url#1{\texttt{#1}}\fi
\expandafter\ifx\csname urlprefix\endcsname\relax\def\urlprefix{URL }\fi
\expandafter\ifx\csname doiprefix\endcsname\relax\def\doiprefix{DOI: }\fi
\providecommand{\bibinfo}[2]{#2}
\providecommand{\eprint}[2][]{\url{#2}}

\bibitem{watt2021correcting}
\bibinfo{author}{Watt-Meyer, O.} \emph{et~al.}
\newblock \bibinfo{journal}{\bibinfo{title}{Correcting weather and climate models by machine learning nudged historical simulations}}.
\newblock {\emph{\JournalTitle{Geophysical Research Letters}}} \textbf{\bibinfo{volume}{48}}, \bibinfo{pages}{e2021GL092555} (\bibinfo{year}{2021}).

\bibitem{kochkov2024neural}
\bibinfo{author}{Kochkov, D.} \emph{et~al.}
\newblock \bibinfo{journal}{\bibinfo{title}{Neural general circulation models for weather and climate}}.
\newblock {\emph{\JournalTitle{Nature}}} \textbf{\bibinfo{volume}{632}}, \bibinfo{pages}{1060--1066} (\bibinfo{year}{2024}).

\bibitem{lu2021learning}
\bibinfo{author}{Lu, L.}, \bibinfo{author}{Jin, P.}, \bibinfo{author}{Pang, G.}, \bibinfo{author}{Zhang, Z.} \& \bibinfo{author}{Karniadakis, G.~E.}
\newblock \bibinfo{journal}{\bibinfo{title}{Learning nonlinear operators via deeponet based on the universal approximation theorem of operators}}.
\newblock {\emph{\JournalTitle{Nature Machine Intelligence}}} \textbf{\bibinfo{volume}{3}}, \bibinfo{pages}{218--229} (\bibinfo{year}{2021}).

\bibitem{li2020fourier}
\bibinfo{author}{Li, Z.} \emph{et~al.}
\newblock \bibinfo{journal}{\bibinfo{title}{Fourier neural operator for parametric partial differential equations}}.
\newblock {\emph{\JournalTitle{arXiv preprint arXiv:2010.08895}}}  (\bibinfo{year}{2020}).

\bibitem{kovachki2021neural}
\bibinfo{author}{Kovachki, N.} \emph{et~al.}
\newblock \bibinfo{journal}{\bibinfo{title}{Neural operator: Learning maps between function spaces}}.
\newblock {\emph{\JournalTitle{arXiv preprint arXiv:2108.08481}}}  (\bibinfo{year}{2021}).

\bibitem{li2020multipole}
\bibinfo{author}{Li, Z.} \emph{et~al.}
\newblock \bibinfo{journal}{\bibinfo{title}{Multipole graph neural operator for parametric partial differential equations}}.
\newblock {\emph{\JournalTitle{Advances in Neural Information Processing Systems}}} \textbf{\bibinfo{volume}{33}}, \bibinfo{pages}{6755--6766} (\bibinfo{year}{2020}).

\bibitem{lam2023learning}
\bibinfo{author}{Lam, R.} \emph{et~al.}
\newblock \bibinfo{journal}{\bibinfo{title}{Learning skillful medium-range global weather forecasting}}.
\newblock {\emph{\JournalTitle{Science}}} \textbf{\bibinfo{volume}{382}}, \bibinfo{pages}{1416--1421} (\bibinfo{year}{2023}).

\bibitem{bi2023accurate}
\bibinfo{author}{Bi, K.} \emph{et~al.}
\newblock \bibinfo{journal}{\bibinfo{title}{Accurate medium-range global weather forecasting with 3d neural networks}}.
\newblock {\emph{\JournalTitle{Nature}}} \textbf{\bibinfo{volume}{619}}, \bibinfo{pages}{533--538} (\bibinfo{year}{2023}).

\bibitem{kurth2023fourcastnet}
\bibinfo{author}{Kurth, T.} \emph{et~al.}
\newblock \bibinfo{title}{Fourcastnet: Accelerating global high-resolution weather forecasting using adaptive fourier neural operators}.
\newblock In \emph{\bibinfo{booktitle}{Proceedings of the platform for advanced scientific computing conference}}, \bibinfo{pages}{1--11} (\bibinfo{year}{2023}).

\bibitem{cciccek20163d}
\bibinfo{author}{{\c{C}}i{\c{c}}ek, {\"O}.}, \bibinfo{author}{Abdulkadir, A.}, \bibinfo{author}{Lienkamp, S.~S.}, \bibinfo{author}{Brox, T.} \& \bibinfo{author}{Ronneberger, O.}
\newblock \bibinfo{title}{3d u-net: learning dense volumetric segmentation from sparse annotation}.
\newblock In \emph{\bibinfo{booktitle}{Medical Image Computing and Computer-Assisted Intervention--MICCAI 2016: 19th International Conference, Athens, Greece, October 17-21, 2016, Proceedings, Part II 19}}, \bibinfo{pages}{424--432} (\bibinfo{organization}{Springer}, \bibinfo{year}{2016}).

\bibitem{Golaz2022E3SMv2}
\bibinfo{author}{Golaz, J.} \emph{et~al.}
\newblock \bibinfo{journal}{\bibinfo{title}{{The DOE E3SM Model Version 2: Overview of the Physical Model and Initial Model Evaluation}}}.
\newblock {\emph{\JournalTitle{Journal of Advances in Modeling Earth Systems}}} \textbf{\bibinfo{volume}{14}}, \bibinfo{pages}{e2022MS003156}, \doiprefix\url{10.1029/2022MS003156} (\bibinfo{year}{2022}).

\bibitem{perez2018film}
\bibinfo{author}{Perez, E.}, \bibinfo{author}{Strub, F.}, \bibinfo{author}{de~Vries, H.}, \bibinfo{author}{Dumoulin, V.} \& \bibinfo{author}{Courville, A.}
\newblock \bibinfo{title}{Film: Visual reasoning with a general conditioning layer}.
\newblock In \emph{\bibinfo{booktitle}{Proceedings of the AAAI Conference on Artificial Intelligence}}, vol.~\bibinfo{volume}{32} (\bibinfo{year}{2018}).

\bibitem{alexeedm}
\bibinfo{author}{Alexeev, D.}
\newblock \bibinfo{journal}{\bibinfo{title}{pytorch-fortran}}.
\newblock {\emph{\JournalTitle{GitHub repository}}}  (\bibinfo{year}{2020}).

\bibitem{hoskins1990stormtracks}
\bibinfo{author}{Hoskins, B.~J.} \& \bibinfo{author}{Valdes, P.~J.}
\newblock \bibinfo{journal}{\bibinfo{title}{On the existence of storm-tracks}}.
\newblock {\emph{\JournalTitle{Journal of the Atmospheric Sciences}}} \textbf{\bibinfo{volume}{47}}, \bibinfo{pages}{1854--1864}, \doiprefix\url{10.1175/1520-0469(1990)047<1854:OTEOST>2.0.CO;2} (\bibinfo{year}{1990}).

\bibitem{minobe2008sst}
\bibinfo{author}{Minobe, S.}, \bibinfo{author}{Kuwano-Yoshida, A.}, \bibinfo{author}{Komori, N.}, \bibinfo{author}{Xie, S.-P.} \& \bibinfo{author}{Small, J.}
\newblock \bibinfo{journal}{\bibinfo{title}{Influence of the gulf stream on the troposphere}}.
\newblock {\emph{\JournalTitle{Nature}}} \textbf{\bibinfo{volume}{452}}, \bibinfo{pages}{206--209}, \doiprefix\url{10.1038/nature06690} (\bibinfo{year}{2008}).

\bibitem{pfahl2015moist}
\bibinfo{author}{Pfahl, S.}, \bibinfo{author}{Madonna, E.}, \bibinfo{author}{Boettcher, M.}, \bibinfo{author}{Joos, H.} \& \bibinfo{author}{Wernli, H.}
\newblock \bibinfo{journal}{\bibinfo{title}{Importance of latent heat release in ascending air streams for atmospheric blocking}}.
\newblock {\emph{\JournalTitle{Nature Geoscience}}} \textbf{\bibinfo{volume}{8}}, \bibinfo{pages}{610--614}, \doiprefix\url{10.1038/ngeo2487} (\bibinfo{year}{2015}).

\bibitem{Santer2000}
\bibinfo{author}{Santer, B.~D.} \emph{et~al.}
\newblock \bibinfo{journal}{\bibinfo{title}{Statistical significance of trends and trend differences in layer‐average atmospheric temperature time series}}.
\newblock {\emph{\JournalTitle{Journal of Geophysical Research: Atmospheres}}} \textbf{\bibinfo{volume}{105}}, \bibinfo{pages}{7337--7356}, \doiprefix\url{10.1029/1999JD901105} (\bibinfo{year}{2000}).

\bibitem{barthel2024debiasing}
\bibinfo{author}{Barthel~Sorensen, B.} \emph{et~al.}
\newblock \bibinfo{journal}{\bibinfo{title}{A non-intrusive machine learning framework for debiasing long-time coarse resolution climate simulations and quantifying rare events statistics}}.
\newblock {\emph{\JournalTitle{arXiv preprint arXiv:2402.18484}}}  (\bibinfo{year}{2024}).

\bibitem{LHeureux2017}
\bibinfo{author}{L'Heureux, M.~L.} \emph{et~al.}
\newblock \bibinfo{journal}{\bibinfo{title}{Observing and predicting the 2015/16 el niño}}.
\newblock {\emph{\JournalTitle{Bulletin of the American Meteorological Society}}} \textbf{\bibinfo{volume}{98}}, \bibinfo{pages}{1363--1382}, \doiprefix\url{10.1175/BAMS-D-16-0009.1} (\bibinfo{year}{2017}).

\bibitem{doecode_10475}
\bibinfo{author}{E3SM~Project, D.}
\newblock \bibinfo{title}{Energy exascale earth system model v1.0}.
\newblock \bibinfo{howpublished}{[Computer Software] \url{https://doi.org/10.11578/E3SM/dc.20180418.36}}, \doiprefix\url{10.11578/E3SM/dc.20180418.36} (\bibinfo{year}{2018}).

\bibitem{Bretherton2022}
\bibinfo{author}{Bretherton, C.~S.} \emph{et~al.}
\newblock \bibinfo{journal}{\bibinfo{title}{Correcting coarse-grid weather and climate models by machine learning from global storm-resolving simulations}}.
\newblock {\emph{\JournalTitle{Journal of Advances in Modeling Earth Systems}}} \textbf{\bibinfo{volume}{14}}, \bibinfo{pages}{e2021MS002794}, \doiprefix\url{10.1029/2021MS002794} (\bibinfo{year}{2022}).

\bibitem{Zhang_et_al:2022}
\bibinfo{author}{Zhang, S.}, \bibinfo{author}{Zhang, K.}, \bibinfo{author}{Wan, H.} \& \bibinfo{author}{Sun, J.}
\newblock \bibinfo{journal}{\bibinfo{title}{Further improvement and evaluation of nudging in the e3sm atmosphere model version 1 (eamv1): simulations of the mean climate, weather events, and anthropogenic aerosol effects}}.
\newblock {\emph{\JournalTitle{Geoscientific Model Development}}} \textbf{\bibinfo{volume}{15}}, \bibinfo{pages}{6787--6816}, \doiprefix\url{10.5194/gmd-15-6787-2022} (\bibinfo{year}{2022}).

\bibitem{Zhang_et_al:2025}
\bibinfo{author}{Zhang, S.}, \bibinfo{author}{Leung, R.} \& \bibinfo{author}{et~al.}
\newblock \bibinfo{title}{Revisiting nudging strategies for machine learning–enabled bias correction in climate models} (\bibinfo{year}{2025}).
\newblock \bibinfo{note}{Manuscript in preparation}.

\bibitem{Gates_et_al:1999}
\bibinfo{author}{Gates, W.~L.} \emph{et~al.}
\newblock \bibinfo{journal}{\bibinfo{title}{An overview of the atmospheric model intercomparison project (amip)}}.
\newblock {\emph{\JournalTitle{Bulletin of the American Meteorological Society}}} \textbf{\bibinfo{volume}{80}}, \bibinfo{pages}{29--55}, \doiprefix\url{10.1175/1520-0477(1999)080<0029:AOOTAM>2.0.CO;2} (\bibinfo{year}{1999}).

\bibitem{Reynolds_2002_SST}
\bibinfo{author}{Reynolds, R.~W.}, \bibinfo{author}{Rayner, N.~A.}, \bibinfo{author}{Smith, T.~M.}, \bibinfo{author}{Stokes, D.~C.} \& \bibinfo{author}{Wang, W.}
\newblock \bibinfo{journal}{\bibinfo{title}{An improved in situ and satellite sst analysis for climate}}.
\newblock {\emph{\JournalTitle{Journal of Climate}}} \textbf{\bibinfo{volume}{15}}, \bibinfo{pages}{1609--1625}, \doiprefix\url{10.1175/1520-0442(2002)015<1609:AIISAS>2.0.CO;2} (\bibinfo{year}{2002}).

\bibitem{Eyring_et_al:2016}
\bibinfo{author}{Eyring, V.}, \bibinfo{author}{Bony, S.}, \bibinfo{author}{Meehl, G.~A.} \emph{et~al.}
\newblock \bibinfo{journal}{\bibinfo{title}{Overview of the coupled model intercomparison project phase 6 (cmip6) experimental design and organization}}.
\newblock {\emph{\JournalTitle{Geoscientific Model Development}}} \textbf{\bibinfo{volume}{9}}, \bibinfo{pages}{1937--1958}, \doiprefix\url{10.5194/gmd-9-1937-2016} (\bibinfo{year}{2016}).

\bibitem{Hoesly_et_al:2018}
\bibinfo{author}{Hoesly, R.~M.}, \bibinfo{author}{Smith, S.~J.}, \bibinfo{author}{Feng, L.} \emph{et~al.}
\newblock \bibinfo{journal}{\bibinfo{title}{Historical (1750–2014) anthropogenic emissions of reactive gases and aerosols from the community emissions data system (ceds)}}.
\newblock {\emph{\JournalTitle{Geoscientific Model Development}}} \textbf{\bibinfo{volume}{11}}, \bibinfo{pages}{369--408}, \doiprefix\url{10.5194/gmd-11-369-2018} (\bibinfo{year}{2018}).

\bibitem{Feng_et_al:2020}
\bibinfo{author}{Feng, Y.}, \bibinfo{author}{Burrows, S.~M.}, \bibinfo{author}{Lin, G.} \emph{et~al.}
\newblock \bibinfo{journal}{\bibinfo{title}{Aerosol–cloud–radiation interactions in the energy exascale earth system model version 1 (e3smv1)}}.
\newblock {\emph{\JournalTitle{Journal of Advances in Modeling Earth Systems}}} \textbf{\bibinfo{volume}{12}}, \doiprefix\url{10.1029/2019MS001851} (\bibinfo{year}{2020}).

\bibitem{szegedy2016rethinking}
\bibinfo{author}{Szegedy, C.}, \bibinfo{author}{Vanhoucke, V.}, \bibinfo{author}{Ioffe, S.}, \bibinfo{author}{Shlens, J.} \& \bibinfo{author}{Wojna, Z.}
\newblock \bibinfo{title}{Rethinking the inception architecture for computer vision}.
\newblock In \emph{\bibinfo{booktitle}{Proceedings of the IEEE conference on computer vision and pattern recognition}}, \bibinfo{pages}{2818--2826} (\bibinfo{year}{2016}).

\bibitem{brenowitz2018prognostic}
\bibinfo{author}{Brenowitz, N.~D.} \& \bibinfo{author}{Bretherton, C.~S.}
\newblock \bibinfo{journal}{\bibinfo{title}{Prognostic validation of a neural network unified physics parameterization}}.
\newblock {\emph{\JournalTitle{Geophysical Research Letters}}} \textbf{\bibinfo{volume}{45}}, \bibinfo{pages}{6289--6298}, \doiprefix\url{10.1029/2018GL078510} (\bibinfo{year}{2018}).

\bibitem{rasp2018deep}
\bibinfo{author}{Rasp, S.}, \bibinfo{author}{Pritchard, M.~S.} \& \bibinfo{author}{Gentine, P.}
\newblock \bibinfo{journal}{\bibinfo{title}{Deep learning to represent subgrid processes in climate models}}.
\newblock {\emph{\JournalTitle{Proceedings of the National Academy of Sciences}}} \textbf{\bibinfo{volume}{115}}, \bibinfo{pages}{9684--9689}, \doiprefix\url{10.1073/pnas.1810286115} (\bibinfo{year}{2018}).

\bibitem{chantry2021machine}
\bibinfo{author}{Chantry, M.}, \bibinfo{author}{Christensen, H.~M.}, \bibinfo{author}{Dueben, P.~D.} \& \bibinfo{author}{Palmer, T.~N.}
\newblock \bibinfo{journal}{\bibinfo{title}{Machine learning emulation of dynamical systems: pitfalls and promise}}.
\newblock {\emph{\JournalTitle{Philosophical Transactions of the Royal Society A}}} \textbf{\bibinfo{volume}{379}}, \bibinfo{pages}{20200083}, \doiprefix\url{10.1098/rsta.2020.0083} (\bibinfo{year}{2021}).

\bibitem{Haarsma2016}
\bibinfo{author}{Haarsma, R.~J.} \emph{et~al.}
\newblock \bibinfo{journal}{\bibinfo{title}{High resolution model intercomparison project (highresmip v1.0) for cmip6}}.
\newblock {\emph{\JournalTitle{Geosci. Model Dev.}}} \textbf{\bibinfo{volume}{9}}, \bibinfo{pages}{4185--4208}, \doiprefix\url{10.5194/gmd-9-4185-2016} (\bibinfo{year}{2016}).

\bibitem{Bryan2019}
\bibinfo{author}{Bryan, F.~O.} \emph{et~al.}
\newblock \bibinfo{journal}{\bibinfo{title}{The ncar cesm high-resolution simulation campaign}}.
\newblock {\emph{\JournalTitle{J. Adv. Model. Earth Syst.}}} \textbf{\bibinfo{volume}{11}}, \bibinfo{pages}{4183--4208}, \doiprefix\url{10.1029/2019MS001763} (\bibinfo{year}{2019}).

\bibitem{Roberts2020}
\bibinfo{author}{Roberts, M.~J.} \emph{et~al.}
\newblock \bibinfo{journal}{\bibinfo{title}{Projecting the benefits of high-resolution global climate modeling}}.
\newblock {\emph{\JournalTitle{Bull. Amer. Meteor. Soc.}}} \textbf{\bibinfo{volume}{101}}, \bibinfo{pages}{E1449--E1476}, \doiprefix\url{10.1175/BAMS-D-19-0309.1} (\bibinfo{year}{2020}).

\bibitem{Balaji2017}
\bibinfo{author}{Balaji, V.} \emph{et~al.}
\newblock \bibinfo{journal}{\bibinfo{title}{Requirements for a global data infrastructure in support of cmip6}}.
\newblock {\emph{\JournalTitle{Geosci. Model Dev.}}} \textbf{\bibinfo{volume}{10}}, \bibinfo{pages}{3775--3791}, \doiprefix\url{10.5194/gmd-10-3775-2017} (\bibinfo{year}{2017}).

\bibitem{Held2019}
\bibinfo{author}{Held, I.~M.} \emph{et~al.}
\newblock \bibinfo{journal}{\bibinfo{title}{Structure and performance of gfdl’s cm4.0 climate model}}.
\newblock {\emph{\JournalTitle{J. Adv. Model. Earth Syst.}}} \textbf{\bibinfo{volume}{11}}, \bibinfo{pages}{3691--3727}, \doiprefix\url{10.1029/2019MS001829} (\bibinfo{year}{2019}).

\bibitem{Bacmeister2014}
\bibinfo{author}{Bacmeister, J.~T.} \emph{et~al.}
\newblock \bibinfo{journal}{\bibinfo{title}{Exploratory high-resolution climate simulations using the community atmosphere model (cam)}}.
\newblock {\emph{\JournalTitle{J. Climate}}} \textbf{\bibinfo{volume}{27}}, \bibinfo{pages}{3073--3099}, \doiprefix\url{10.1175/JCLI-D-13-00387.1} (\bibinfo{year}{2014}).

\bibitem{Caldwell2019}
\bibinfo{author}{Caldwell, P.~M.} \emph{et~al.}
\newblock \bibinfo{journal}{\bibinfo{title}{The doe e3sm coupled model version 1: Overview and evaluation at standard and high resolution}}.
\newblock {\emph{\JournalTitle{J. Adv. Model. Earth Syst.}}} \textbf{\bibinfo{volume}{11}}, \bibinfo{pages}{4095--4146}, \doiprefix\url{10.1029/2019MS001870} (\bibinfo{year}{2019}).

\end{thebibliography}
\clearpage

\section{Theoretical Analysis}
\noindent \textbf{Lemma 2.3.1:} Let $K_{i}:\mathbb{R}^{n}\rightarrow \mathbb{R}^{m_{i}}$, $i=1,2,\cdot\cdot\cdot r$ be the linear maps, and let $\Gamma(K_{i}^{'})\subseteq\mathbb{R}^{n}$ be the row space of $K_{i}$, defined by
\begin{align}
    K &:\mathbb{R}^{n}\rightarrow \mathbb{R}^{m_{1}}\oplus\mathbb{R}^{m_{2}}\oplus \cdot \cdot \cdot \oplus \mathbb{R}^{m_{r}} \\
    K(v) &= (K_1(v),K_2(v),\cdots ,K_r(v)),
\end{align}then rank of $K$ is given by
\begin{equation}
    \text{rank}(K) = \text{dim}\left( \sum_{i=1}^{r}\Gamma(K_{i}^{'}) \right), 
\end{equation}and is monotonic. Furthermore, $K$ is injective iff rank($K$) = n and the left inverse is given by the Moore Penrose pseudoinverse defined by
\begin{equation}
    K^{+} = (K^{'}K)^{-1}K^{'},  
\end{equation}such that $K^{+}K = I_{n}$. \newline

\noindent \textbf{Sketch of proof:}\newline

\noindent Let $A_{i}$ be a $m_{i} \times n$ matrix, such that , $K:\mathbb{R}^{n}\rightarrow \mathbb{R}^{m_{1}}\oplus\mathbb{R}^{m_{2}}\oplus \cdot \cdot \cdot \oplus \mathbb{R}^{m_{r}}$ is represented by \begin{equation}
    A = \begin{pmatrix}
A_1 \\
A_2 \\
\vdots \\
A_r
\end{pmatrix},
\quad
m = \sum_{i=1}^{r} m_i.
\end{equation} Hence, by definition \begin{equation}
    \text{rank}(K) = \text{dim}(\text{row}(A))
\end{equation} and \begin{equation}
    \text{row}(A) = \text{span}\left( \bigcup_{i=1}^{r}\text{row}A_{i} \right) = \sum_{i=1}^{r}\Gamma(A_{i}).
\end{equation}Since, row$(A_{i})=\Gamma(K_{i}^{'})$, therefore \begin{equation}
    \text{rank}(K) = \text{dim}\left(\sum_{i=1}^{r}\Gamma(K_{i}^{'}) \right).
\end{equation}For $i \in \mathbb{N}$, \begin{equation}
    \Gamma(K_{i}^{'}) \subseteq \sum_{j=1}^{r}\Gamma(K_{j}^{'}).
\end{equation}Therefore \begin{align}
    \text{dim}\left( \sum_{j=1}^{r}\Gamma(K_{j}^{'})\right) &\geq \text{dim}\Gamma(K_{i}^{'}) \\
    & = \text{rank}(K_{i}^{'}) \\
    &= \text{rank}(K_{i}).
\end{align}Therefore by considering maximum $i$, we have \begin{equation}
    \text{rank}(K) \geq \max_{1\leq i \leq r}\text{rank}(K_{i}).
\end{equation}Now let us assume \begin{equation}
    \sum_{i=1}^{r}\Gamma(K_{i}^{'}) = \mathbb{R}^{n}.
\end{equation}Then,
\begin{align}
    \text{dim}\left( \sum_{i=1}^{r}\Gamma(K_{i}^{'}) \right) & = n \\
    \text{rank}(K) & = n.
\end{align}Therefore, A is of full rank and is invertible. Let $K^{+}=(A^{'}A)^{-1}A^{'}$, then
\begin{equation}
    K^{+}K = (A^{'}A)^{-1}A^{'}A = I_{n}.
\end{equation}Therefore, $K$ is injective and $K^{+}$ is a left inverse.

\section{Error and Quality Metrics}

Let $\{y_i\}_{i=1}^n$ be the ground‐truth values, $\{\hat y_i\}_{i=1}^n$ the corresponding predictions, and define the error $e_i = \hat y_i - y_i$ and the sample mean
\[
\bar y = \frac{1}{n}\sum_{i=1}^n y_i.
\]
We then consider the following error metrics for comparison as defined below.
\begin{description}
  \item[\bfseries Mean Squared Error (MSE)] 
    Quantifies the average squared discrepancy between prediction and truth:
    \[
      \mathrm{MSE}
      = \frac{1}{n} \sum_{i=1}^n e_i^2.
    \]

  \item[\bfseries Root Mean Squared Error (RMSE)]
    The square root of the MSE, restoring original units:
    \[
      \mathrm{RMSE}
      = \sqrt{\mathrm{MSE}}
      = \sqrt{\frac{1}{n}\sum_{i=1}^n e_i^2}.
    \]

  \item[\bfseries Mean Absolute Error (MAE)]
    Measures the average magnitude of the errors:
    \[
      \mathrm{MAE}
      = \frac{1}{n}\sum_{i=1}^n \lvert e_i\rvert.
    \]

  \item[\bfseries Peak Signal‐to‐Noise Ratio (PSNR)]
    Expresses the MSE on a logarithmic scale relative to the dynamic range:
    \[
      \mathrm{PSNR}
      = 10\log_{10}\!\Bigl(\frac{L_{\max}^2}{\mathrm{MSE}}\Bigr),
      \quad
      L_{\max} = \max_i y_i - \min_i y_i.
    \]

  \item[\bfseries Bias]
    The mean error, indicating systematic over‐ or under‐prediction:
    \[
      \mathrm{Bias}
      = \frac{1}{n}\sum_{i=1}^n e_i.
    \]

  \item[\bfseries Standard Error of the Estimate]
    The standard deviation of the residuals around the bias:
    \[
      \mathrm{StdError}
      = \sqrt{
          \frac{1}{n}\sum_{i=1}^n\bigl(e_i - \mathrm{Bias}\bigr)^2
        }.
    \]

  \item[\bfseries Coefficient of Determination ($R^2$)]
    Fraction of variance in $y$ explained by the model:
    \[
      R^2
      = 1 - \frac{\sum_{i=1}^n e_i^2}{\sum_{i=1}^n (y_i - \bar y)^2}.
    \]

  \item[\bfseries Coefficient of Variation of the Error (CV error)]
    The RMSE normalized by the mean of the truth:
    \[
      \mathrm{CV}
      = \frac{\mathrm{RMSE}}{\bar y}.
    \]

  \item[\bfseries Structural Similarity Index (SSIM)]
    A perceptual metric for image‐like fields, comparing local means, variances and covariance:
    \[
      \mathrm{SSIM}(x,y)
      = \frac{(2\mu_x\mu_y + C_1)\,(2\sigma_{xy}+C_2)}
             {(\mu_x^2+\mu_y^2 + C_1)\,(\sigma_x^2+\sigma_y^2 + C_2)},
    \]
    where $\mu_x,\mu_y$ and $\sigma_x^2,\sigma_y^2,\sigma_{xy}$ are local statistics and $C_1,C_2$ are small constants.
\end{description}

\section{PyTorch-FORTRAN package}

\begin{itemize}
    \item Load and use PyTorch-trained models on all the processors inside Fortran programs.
    \item Perform inference  directly from FORTRAN by loading traced model in the PyTorch.
    \item Use GPU acceleration with minimal overhead (via CUDA/OpenACC).
    \item Avoid writing glue code in C++/Python manually.
\end{itemize}
The steps for utilizing this package are shown below
\begin{enumerate}
    \item Model is trained in Python
    \begin{itemize}
        \item Define and train your model using PyTorch in Python.
        \item Save the model using torch.jit.trace() or torch.jit.script() to create a TorchScript module.
    \end{itemize}
\item Model is loaded in Fortran
Using the pytorch-fortran API( Load the saved TorchScript .pt model in your Fortran code.)

\item Create and manage data structure represented using traceable tensors from Fortran. Call the model with input tensors. Retrieve outputs as native Fortran arrays ready to be used directly for subsequent computation.

\item Data exchange between Fortran and PyTorch
Input/output arrays in Fortran are copied (or zero-copied in GPU mode) to PyTorch tensors.
The binding handles shape and memory alignment conversions. On GPU: OpenACC directives are used to avoid extra memory copies by sharing GPU pointers.
\end{enumerate}

\section{Additional Offline results} 

\begin{figure}[!ht]
    \centering
    \includegraphics[width=0.8\linewidth]{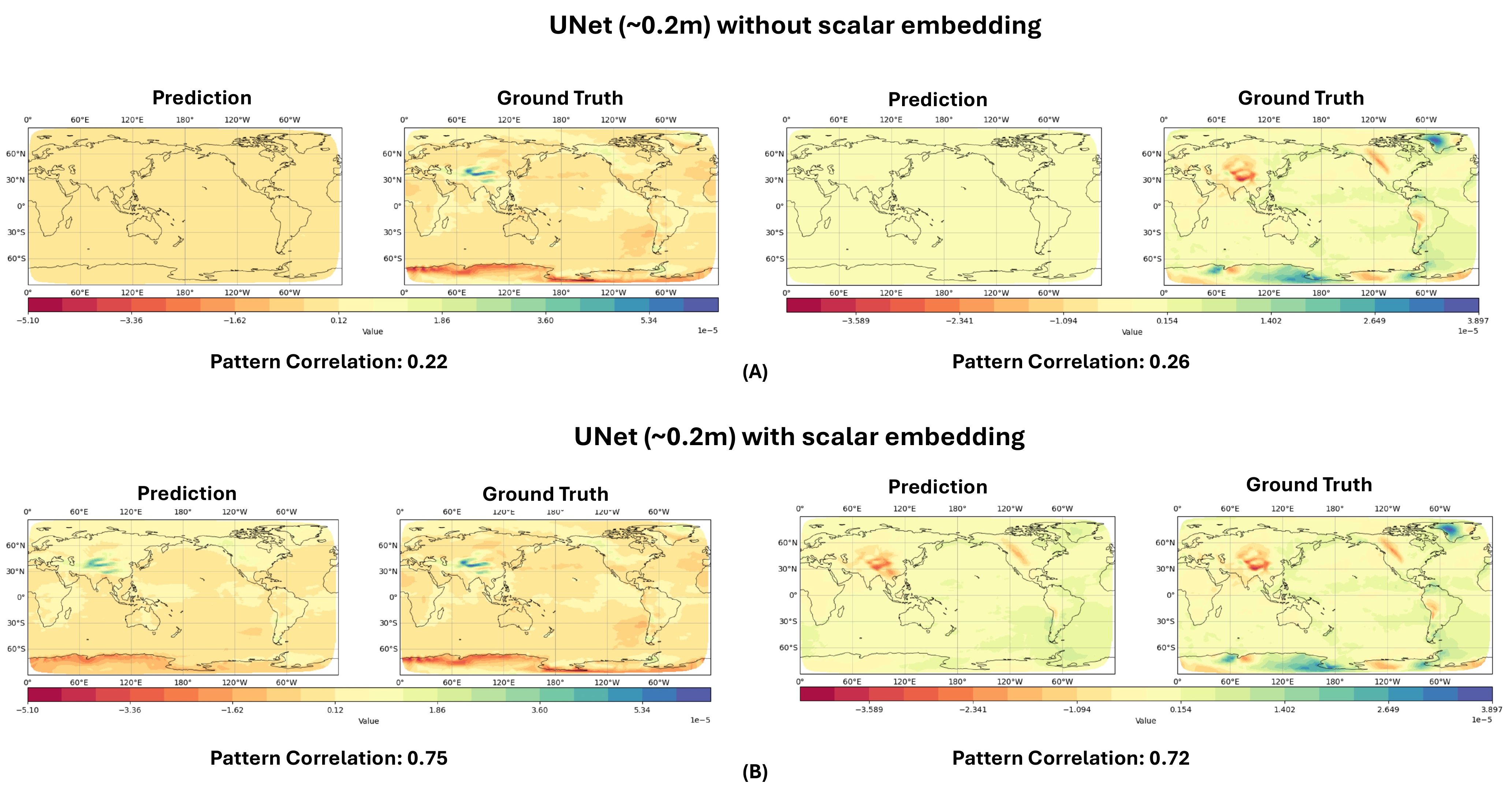}
    \caption{\textbf{Effect of scalar embedding on machine learning nudging tendencies.}
    Show are results from the baseline UNet model ($\sim$0.2\,M parameters) without (panels in row~A) and with (panels in row-B) scalar embedding, respectively. The left and right columns correspond to the annual-mean zonal (U) and meridional (V) nudging tendencies (m\,s$^{-1}$), averaged across all vertical layers during the test year 2015. The inclusion of scalar embedding substantially enhances the spatial correspondence with the reference fields, as reflected by higher pattern correlation coefficients.}
    \label{fig:scalar}
\end{figure}

\begin{table}[!ht]
\centering
\resizebox{\textwidth}{!}{%
\begin{tabular}{lccc}
\toprule
\textbf{Architecture Component} & \textbf{UNet} & \textbf{IUNet} & \textbf{M\&M} \\
\midrule
Encoder & Standard double convolution (DoubleConv) & Inception-style encoder & Standard double convolution (DoubleConv) \\
Downsampling & Max pooling & Max pooling & Max pooling \\
Decoder & Transposed convolution only & Inception-style decoder & Transposed convolution + PixelShuffle + Inception-based upsampling \\
Feature Fusion Strategy & Skip connection + concatenation + convolution & Same as UNet & Same as IUNet + refinement layer (DoubleConv) \\
Refinement Layer & Not explicitly defined & Not explicitly defined & Explicit refinement using DoubleConv after fusion \\
Receptive Field & Local & Expanded via Inception encoder & Expanded via decoder structure \\
Decoder Expressivity & Moderate & Moderate & High (multi-path fusion) \\
\bottomrule
\end{tabular}%
}
\caption{Summary of key architectural components in the UNet, IUNet, and M\&M machine learning models.}
\label{tab:model_comparison}
\end{table}

\begin{figure}[!ht]
    \centering
    \includegraphics[width=\linewidth]{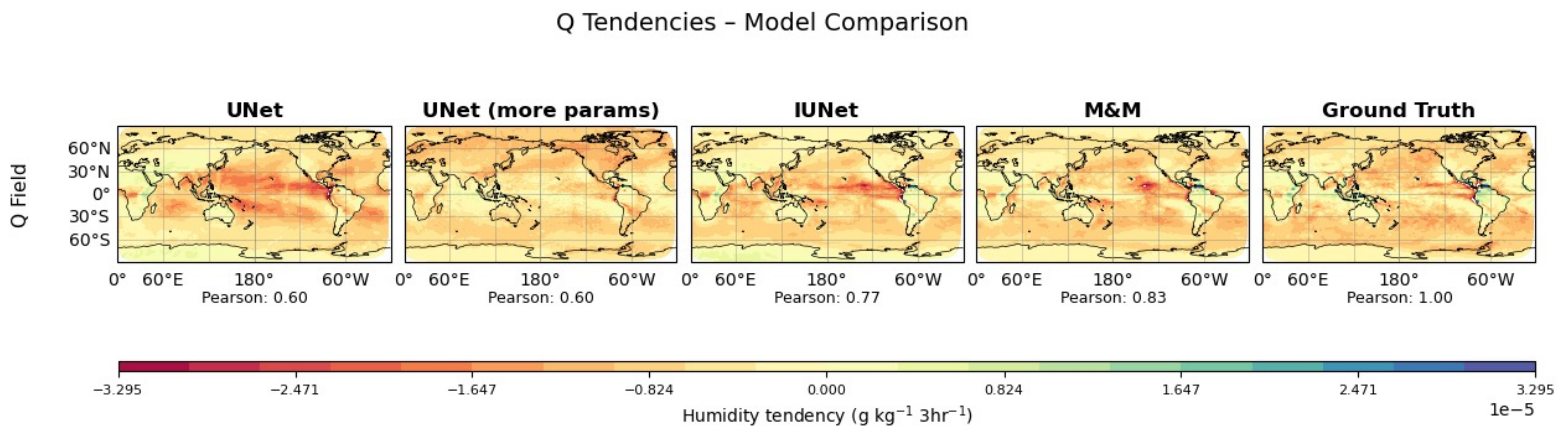}
    \caption{Results showing the arithmetic mean nudging tendency for Q-Humidity (g/kg) across the vertical layers and time over the test year 2015.}
    \label{fig:QNdg_mean}
\end{figure}
\begin{table}[htbp]
\centering
\caption{Performance comparison for $Q$ (Humidity) of different deep learning models. Best entries per metric in $\mathbf{bold}$.}
\label{tab:metric_comparison_q_norf}
\scriptsize
\begin{tabular}{lcccc}
\toprule
\textbf{Metric}         & \textbf{UNet}          & \textbf{UNet (more params)} & \textbf{IUNet}                        & \textbf{M\&M}           \\
\midrule
Parameters              & $\mathbf{\sim0.2\,M}$ & $\sim6.7\,M$                 & $\sim6.7\,M$                          & $\sim6.7\,M$            \\

MSE                     & $1.36885\times10^{-4}$ & $1.31689\times10^{-4}$      & $\mathbf{1.08245\times10^{-4}}$       & $1.10833\times10^{-4}$  \\

RMSE                    & $1.16998\times10^{-2}$ & $1.14756\times10^{-2}$      & $\mathbf{1.04041\times10^{-2}}$       & $1.05277\times10^{-2}$  \\

MAE                     & $4.26589\times10^{-3}$ & $4.19887\times10^{-3}$      & $\mathbf{3.75503\times10^{-3}}$       & $3.76861\times10^{-3}$  \\

PSNR               & $38.64$                & $38.80$                     & $\mathbf{39.66}$                      & $39.55$                 \\

Bias                    & $-4.57844\times10^{-5}$ & $\mathbf{-2.00730\times10^{-5}}$ & $8.07843\times10^{-5}$           & $6.03971\times10^{-5}$  \\

StdError                & $1.16997\times10^{-2}$ & $1.14756\times10^{-2}$      & $\mathbf{1.04037\times10^{-2}}$       & $1.05275\times10^{-2}$  \\

Global $R^2$            & $0.072$                & $0.108$                     & $\mathbf{0.266}$                      & $0.249$                 \\

CV error                & $0.0213$               & $0.0209$                    & $\mathbf{0.0189}$                     & $0.0191$                \\

SSIM                    & $0.818794$             & $0.826877$                  & $\mathbf{0.851614}$                   & $0.849546$              \\
\bottomrule
\end{tabular}
\end{table}

Figure \ref{fig:QNdg_mean} and Table \ref{tab:Q} compare the results for the prediction of the nudging tendency for humidity (Q) for the test year 2015. The UNet with less parameters ($\sim$0.2M) yields a pattern correlation of 0.60, which still remains similar when scaled to $\sim$6.7M parameters. IUNet delivers a pattern correlation ($\sim$0.77) and outperforms all others on every metric. The M\&M model also improves a lot over the UNet and is close to IUNet with the highest pattern correlation of 0.83, and trails slightly behind IUNet over all the metrics. M\&M is good at capturing the broad amplitude distribution and tail extremes of the tendency for U and V. Humidity, however, exhibits a more fragmented fine-scale spatial structure with a tighter dynamic range, so the inception-based multi-path filters in IUNet turn out to be better at resolving those patterns. 
\begin{figure}[!ht]
    \centering
    \includegraphics[width=0.8\linewidth]{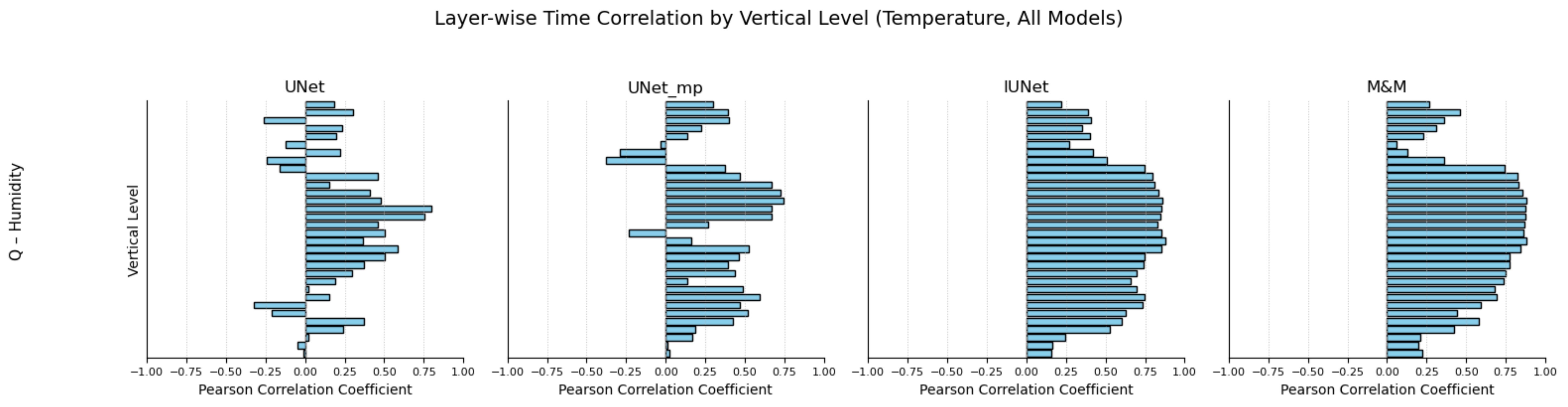}
    \caption{Results showing the time correlation across different vertical layers for Q-Humidity over the test year 2015.}
    \label{fig:CV1}
\end{figure} Figure \ref{fig:CV1} shows the temporal fidelity of the models for humidity Q nudging tendency, for the vertical levels between 26 and 58 for the test year 2015. The UNet with less parameters ($\sim$0.2M) achieves coherence only at a few mid-levels and falls below zero for both lower- and upper-levels. Scaling the UNet to $\sim$6.7M parameters raises its correlation to 0.65 in the mid-levels and reduces most of the negative correlations, but the model still struggles at the near-surface and upper layers. IUNet with the same number of parameters improves the temporal fidelity and sustains a correlation above 0.7 for levels from 30 through 52 and has peak values of 0.9 near level 37-42 and only reduces to $\sim$0.3 at some of the stratospheric levels. The M\&M with the same parameters shows a nearly identical profile, maintaining a high correlation of above 0.6 for most of the levels and peaking at 0.85 near the mid troposphere, and has no negative values. These results show that both IUNet and M\&M yield better temporal coherence for humidity bias corrections. 
\begin{figure}[!ht]
    \centering
    \includegraphics[width=0.6\linewidth]{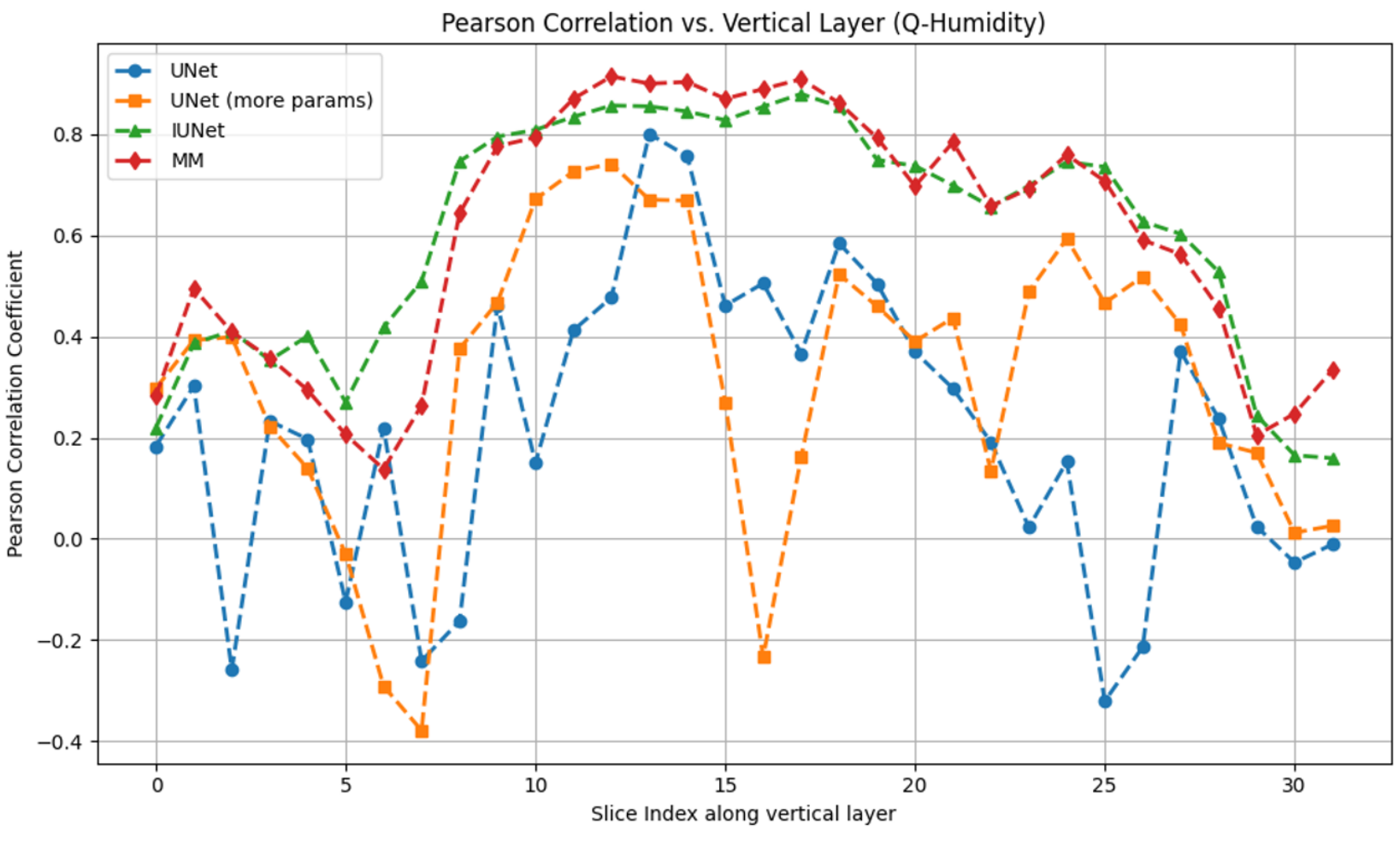}
    \caption{Results showing the correlation across different vertical layers for Q-Humidity over the test year 2015.}
    \label{fig:CV2}
\end{figure}Figure \ref{fig:CV2} shows the layer wise mean Pearson correlation over all time for the test year 2015. The UNet with less parameters displays negative correlation at many of low and high altitude level and has a positive value only for the mid-troposphere level. Scaling the UNet to $\sim$6.7M parameters raises the positive correlation over the mid levels, and reduces the negative correlation. IUNet achieves a high correlation of above 0.7 for levels from 8 through 24 and never falls below 0.1 at any of the other levels. M\&M delivers the most uniform high correlation across the levels and a peak value of 0.9 for mid-levels, demonstrating the effective capture of the vertical structure of humidity tendencies.

Overall, for the Q nudging tendencies, both IUNet and M\&M outperform the UNet variants. IUNet achieves the highest spatial agreement and lowest reconstruction error, whereas M\&M has the highest temporal fidelity with strong spatial correlation.

\begin{figure}[!ht]
    \centering
    \includegraphics[width=\linewidth]{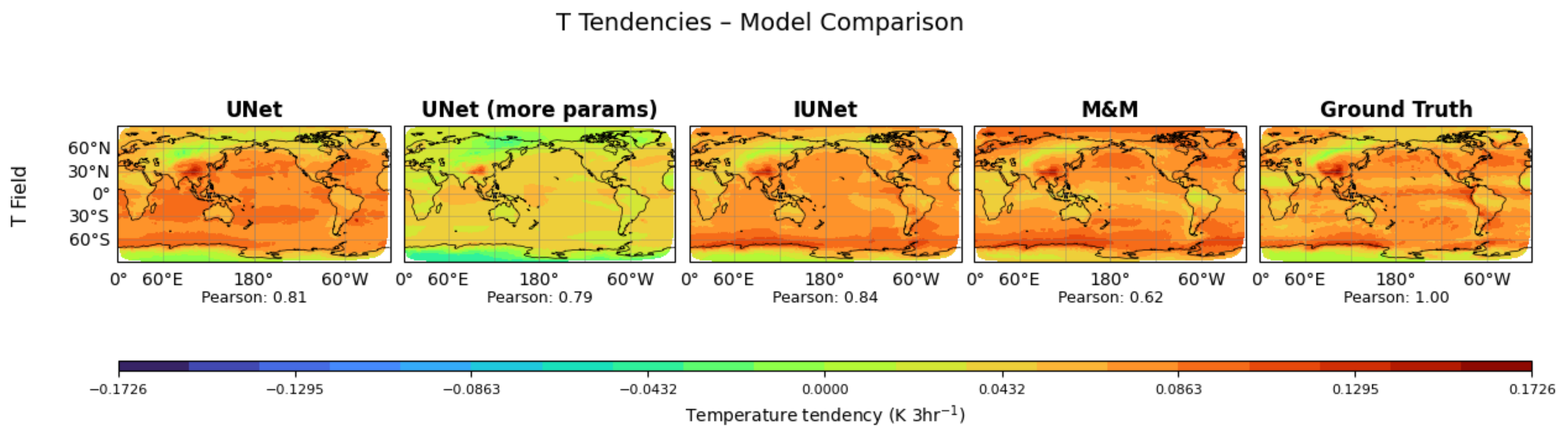}
    \caption{Results showing the arithmetic mean nudging tendency for T-Temperature (K) across the vertical layers and time over the test year 2015.}
    \label{fig:TNdg_mean}
\end{figure}
\begin{figure}[!ht]
    \centering
    \includegraphics[width=\linewidth]{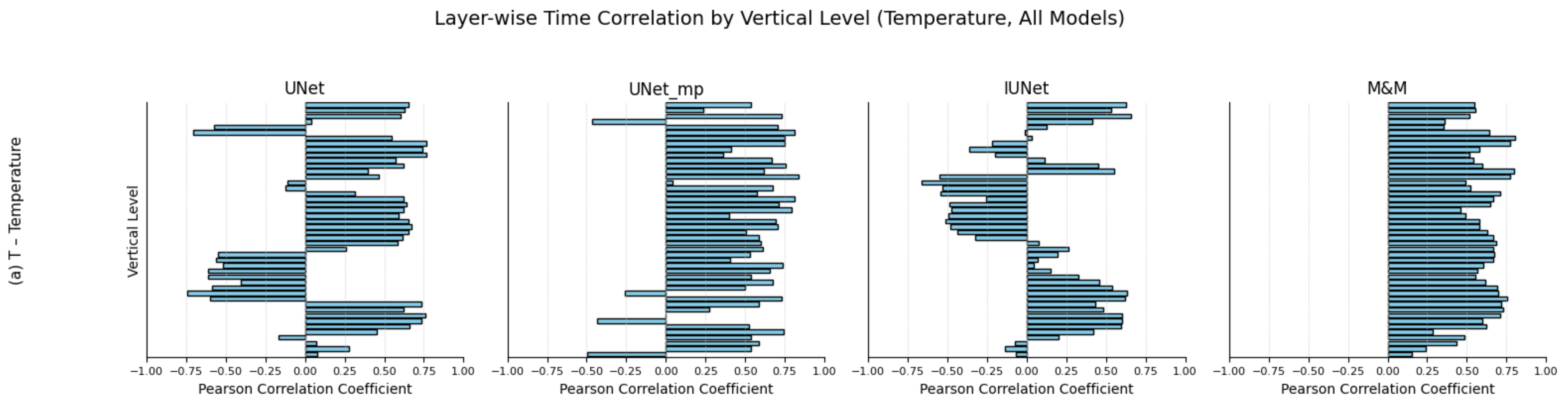}
    \caption{Results showing the time correlation across different vertical layers for T (Temperature) over the test year 2015.}
    \label{fig:CV3}
\end{figure}

\begin{table}[htbp]
\centering
\caption{Performance comparison for $T$ (Temperature) nudging tendency for different models. Best entries per metric in $\mathbf{bold}$.}
\label{tab:transposed_performance_latest}
\scriptsize
\begin{tabular}{lcccc}
\toprule
\textbf{Metric}         & \textbf{UNet}         & \textbf{UNet (more params)} & \textbf{IUNet}       & \textbf{M\&M}         \\
\midrule
Parameters               & $\mathbf{\sim0.2\,\mathrm{M}}$ & $\sim6.7\,\mathrm{M}$        & $\sim6.7\,\mathrm{M}$ & $\sim6.7\,\mathrm{M}$ \\

MSE                & $1.249\times10^{-4}$ & $1.274\times10^{-4}$        & $1.171\times10^{-4}$ & $\mathbf{1.142\times10^{-4}}$ \\

RMSE                & $1.12\times10^{-2}$  & $1.13\times10^{-2}$         & $1.08\times10^{-2}$  & $\mathbf{1.07\times10^{-2}}$   \\

MAE                & $6.83\times10^{-3}$  & $7.04\times10^{-3}$         & $6.56\times10^{-3}$  & $\mathbf{6.54\times10^{-3}}$ \\

PSNR                & $39.03$              & $38.95$                     & $39.32$              & $\mathbf{39.43}$               \\

Bias               & $2.89\times10^{-4}$  & $-1.49\times10^{-3}$        & $\mathbf{7.37\times10^{-5}}$ & $8.60\times10^{-4}$    \\

Std. Error         & $1.12\times10^{-2}$  & $1.12\times10^{-2}$         & $1.08\times10^{-2}$  & $\mathbf{1.06\times10^{-2}}$  \\

Global $R^2$             & $0.110$              & $0.092$                     & $0.166$              & $\mathbf{0.187}$               \\

CV error                 & $0.024$              & $0.024$                     & $\mathbf{0.023}$     & $\mathbf{0.023}$              \\

SSIM                     & $0.572$              & $0.553$                     & $\mathbf{0.610}$     & $0.603$                       \\
\bottomrule
\end{tabular}
\label{S-tab:T}
\end{table}
Figure \ref{fig:TNdg_mean} and Table \ref{tab:T} show that the UNet with fewer parameters ($\sim$0.2M) already reproduces much of the time-mean large-scale structure of the temperature nudging tendency, with a Pearson pattern correlation of 0.81. However, it still has the largest MSE and RMSE among the models and only a modest global $R^{2}\approx 0.11$. Increasing the capacity to $\sim$6.7M parameters does not systematically improve these bulk errors: the wider UNet attains a very similar spatial pattern (Pearson $\approx 0.79$) but slightly worse MSE, RMSE, and $R^{2}$, and introduces more small-scale artifacts. IUNet, with the same number of parameters, provides the sharpest structural agreement with the reference, yielding the highest pattern correlation (0.84), the best SSIM, and the smallest bias while also reducing the error metrics relative to both UNets. M\&M, again with $\sim$6.7M parameters, is the most accurate in a pointwise sense, achieving the lowest MSE and RMSE and the highest PSNR and global $R^{2}\approx 0.19$, but at the cost of a lower large-scale pattern correlation (0.62) and slightly less coherent anomalies in some regions. Thus, from a purely spatial, time-mean perspective, IUNet offers the best compromise between structural fidelity and error magnitude, while M\&M prioritizes local accuracy; the vertical and temporal behavior of these models is examined separately in Figure \ref{fig:CV3}.

 Figure \ref{fig:CV3} shows the temporal fidelity in reproducing nudging tendency for temperature for the model levels 12-58 for the test year 2015. The UNet with less parameters shows highly oscillatory behavior and has a positive correlation between 0.5 and 0.6  for the mid troposphere and a negative correlation for both lower and upper levels, indicating severe temporal misalignment. Scaling the UNet to $\sim$6.7M parameters reduces the worst negative values and raises the mid-level correlation to up to 0.75, and struggles at the top and bottom levels. IUNet with the same number of parameters also has some oscillatory behavior, but has higher correlation than the UNet with less parameters, but worse than the UNet with the same number of parameters.  M\&M with the same number of parameters delivers the most consistent time correlations, maintaining values above 0.6 across almost the entire column and peaking near 0.85 in the upper troposphere. This shows M\&M's ability to capture the full vertical structure of the variability.
\begin{figure}[!ht]
    \centering
    \includegraphics[width=0.8\linewidth]{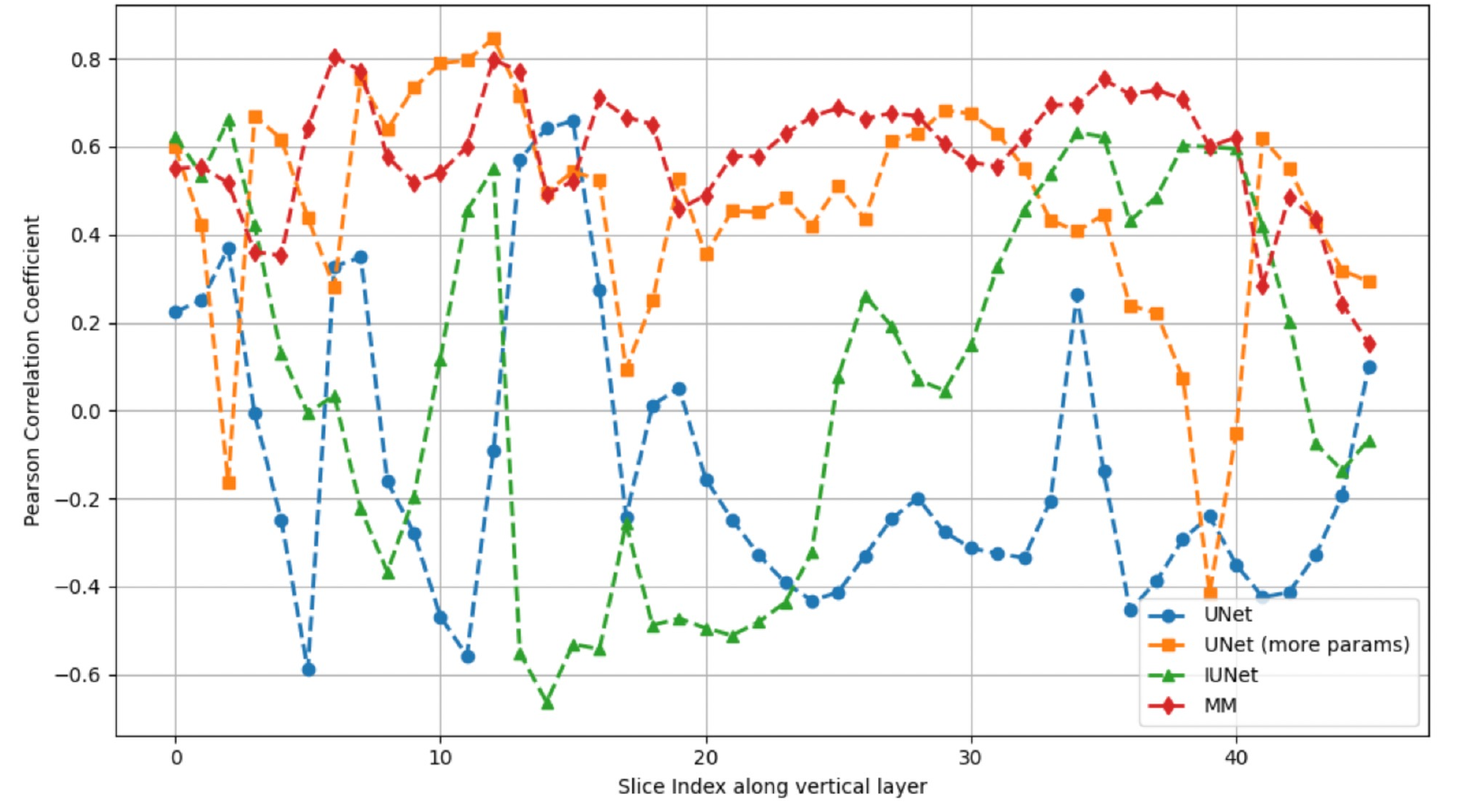}
    \caption{Results showing the correlation across different vertical layers for T-temperature over the test year 2015.}
    \label{fig:CV4}
\end{figure}Figure \ref{fig:CV4} shows the layer-wise Pearson correlation for the predicted T nudging tendency. The UNet with less parameters shows negative correlation at many levels. The UNet with $\sim$6.7M parameters has a much more stable behavior, reduces the negative correlation, and achieves a peak value of 0.8 around levels between 10-12. IUNet has a strong correlation of greater than 0.6 through much of the lower and mid troposphere, but has negative correlations in upper layers. M\&M maintains the most uniform high correlations across all the vertical levels, demonstrating its robust temporal tracking of the tendencies.

\subsection*{Discussion on Deep Learning Versus Traditional ML (Random Forest)} 
Deep learning models such as UNet, IUNet, and the M\&M network learn rich, hierarchical representations by stacking convolutional layers that exploit both local and global spatial context.  They are trained end-to-end to minimize a loss (e.g., \ MSE), automatically discovering features that capture complex non-linear relationships in the data.  In contrast, a random forest builds an ensemble of decision trees by greedily partitioning the input feature space along axis-aligned splits.  It treats each input dimension largely independently and cannot natively capture spatial correlations beyond what can be encoded in engineered features. Since our bias-correction task involves high-resolution fields with multi-scale structure, we expect convolutional networks to outperform random forests:  
\begin{itemize}
  \item \textbf{Spatial inductive bias:}  Convolutions enforce locality and parameter sharing, yielding far fewer parameters than a fully connected tree on every grid cell.
  \item \textbf{Feature learning:}  deep networks learn specialized filters for edges, textures, and global patterns, whereas random forests rely on manually chosen features or flattened inputs.
  \item \textbf{End-to-end optimization:}  All layers in a neural network are tuned jointly to minimize prediction error; tree ensembles optimize splits greedily and independently in each tree.
\end{itemize}
\paragraph{Deep learning}
\begin{itemize}
  \item \emph{Advantages:}  
    \begin{enumerate}
      \item Can model highly non-linear, hierarchical features directly from raw data.  
      \item Captures both local and global structures via convolution and multi-scale architectures.  
      \item Scales to very large datasets, often improving with more data.
    \end{enumerate}
  \item \emph{Limitations:}  
    \begin{enumerate}
      \item Requires large amounts of labeled data and significant computational resources (GPU/TPU).  
      \item Sensitive to hyperparameter choices (learning rate, architecture, regularization).  
      \item Risk of overfitting when data are scarce or noise‐dominated.
    \end{enumerate}
\end{itemize}

\paragraph{Random forest}
\begin{itemize}
  \item \emph{Advantages:}  
    \begin{enumerate}
      \item Fast to train and tune; few hyperparameters (number of trees, depth).  
      \item Robust to overfitting on tabular data; provides feature‐importance measures.  
      \item Low computational and memory footprint compared to deep networks.
    \end{enumerate}
  \item \emph{Limitations:}  
    \begin{enumerate}
      \item Lacks an inherent mechanism for capturing spatial context in high-dimensional fields.  
      \item Performance plateaus on very complex, structured tasks.  
      \item Requires engineered features if raw inputs are not already informative.
    \end{enumerate}
\end{itemize}

Table~\ref{tab:metric_comparison}  shows that all deep models substantially outperform the random forest baseline:

\begin{itemize}
  \item \textbf{Error reduction:}  
    The best deep model (IUNet) achieves $\mathrm{MSE}=1.08245\times10^{-4}$ vs.\ $1.43304\times10^{-4}$ (RF),  
    $\mathrm{RMSE}=1.04041\times10^{-2}$ vs.\ $1.19710\times10^{-2}$,  
    and $\mathrm{MAE}=3.75503\times10^{-3}$ vs.\ $4.22903\times10^{-3}$.  
  \item \textbf{Statistical fit:}  
    Global $R^2$ climbs from $0.029$ (RF) to $0.266$ (IUNet),  
    and CV error falls from $0.0218$ to $0.0189$.  
  \item \textbf{Information‐theoretic gain:}  
    NMI increases from $0.029$ to $0.188$, indicating stronger mutual dependence between predictions and true fields.
\end{itemize}

\textbf{Power Spectrum:} Given a 2D spatial field $ u_t(x, z) \in \mathbb{R}^{N_z \times N_x} $, the radial power spectrum is computed as follows:

\begin{align*}
\hat{u}_t(k_x, k_z) &= \mathbb{F}_{2D}[u_t(x, z)] \\
P_t(k_x, k_z) &= |\hat{u}_t(k_x, k_z)|^2 \\
\mathbb{P}_t(k_i) &= \frac{1}{N_i} \sum_{(k_x, k_z) \in \text{bin } i} P_t(k_x, k_z) \\
\bar{\mathbb{P}}(k_i) &= \frac{1}{N_t} \sum_{t=1}^{N_t} \mathbb{P}_t(k_i)
\end{align*}

where $ k_i = \sqrt{k_x^2 + k_z^2} $ denotes the radial wavenumber bins.

\begin{figure}[!ht]
    \centering
    \includegraphics[width=0.7\linewidth]{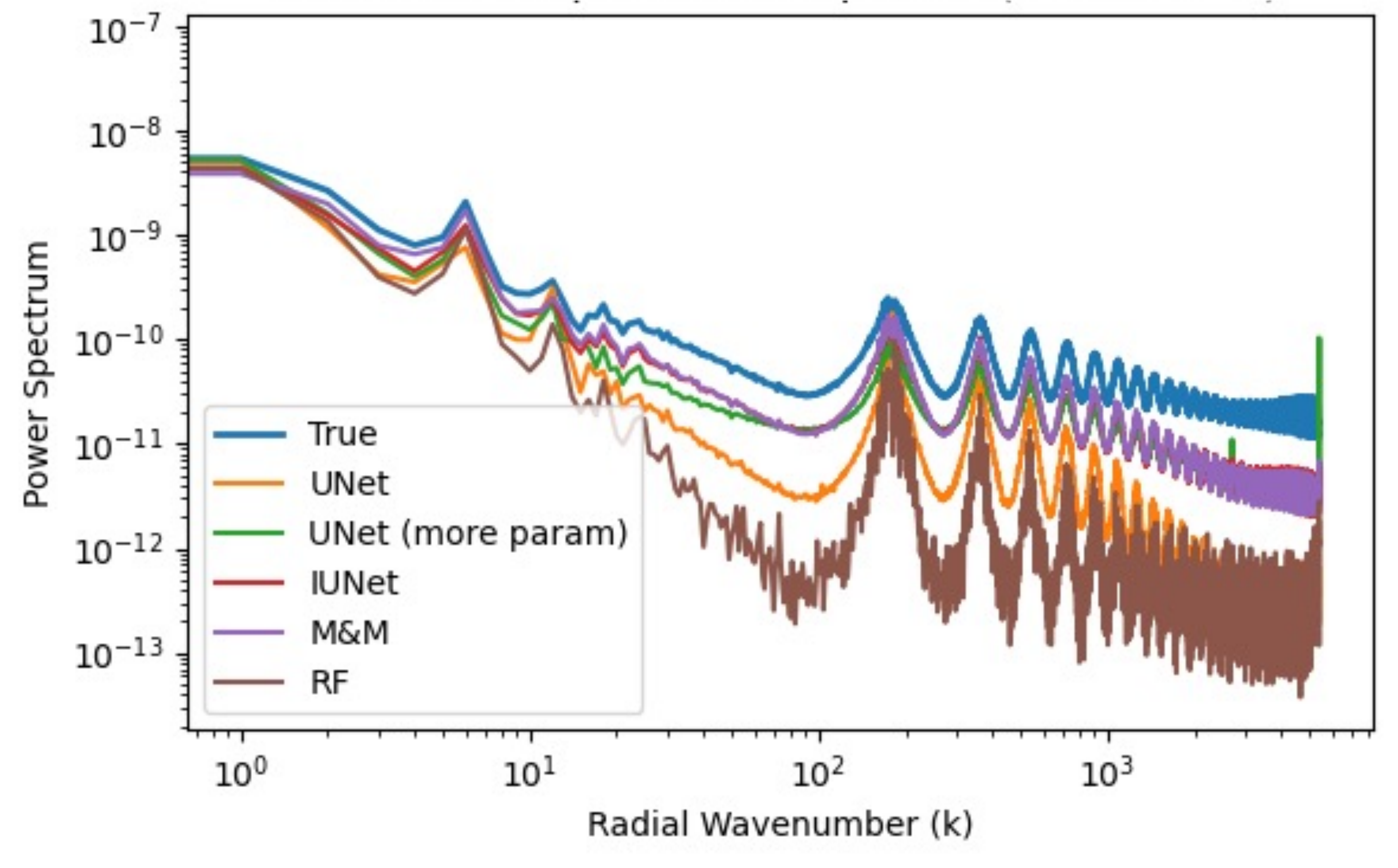}
    \caption{Figure comparing the power spectrum vs radial wave number of different methodologies for the Q (humidity) nudging tendency for the test year 2015.}
    \label{fig:PS}
\end{figure}

The random forest does exhibit a slightly smaller bias magnitude ($1.32\times10^{-11}$) than most deep networks, but this comes at the cost of far higher variance and poor overall fit.  These results confirm that deep convolutional architectures are better suited for recovering fine-scale spatial patterns in our bias‐correction application, while random forests are more appropriate for lower-dimensional or tabular regression tasks. Figure \ref{fig:PS} compares the radial power spectra of the true bias field against predictions from UNet, UNet (more parameters), IUNet, M \& M, and the random‐forest (RF) baseline. All convolutional models capture the large‐scale (low $k$) power very accurately, clustering tightly around the true spectrum up to $k\simeq10$. Beyond this mid‐range, the vanilla UNet begins to under-predict energy, and the RF spectrum decays most rapidly, indicating a failure to reconstruct fine‐scale variability. Adding parameters to the UNet partially restores mid‐range power, but the Inception‐style IUNet and the multi‐scale M \& M network deliver the best fidelity: their spectra remain within one order of magnitude of truth all the way into the dissipative (high $k$) regime. In contrast, RF loses two to three orders of magnitude by $k\simeq 10^2$. This demonstrates that convolutional backbones with multi‐scale feature extraction are substantially more capable of modelling high‐wavenumber corrections than tree-based regressors.

\begin{table}[htbp]
\centering
\caption{Performance comparison for $Q$ (Humidity) of different models with Random Forest. Best entries per metric in $\mathbf{bold}$.}
\label{tab:metric_comparison_q}
\scriptsize
\begin{tabular}{lccccc}
\toprule
\textbf{Metric}         & \textbf{UNet}          & \textbf{UNet (more params)} & \textbf{IUNet}                        & \textbf{M\&M}           & \textbf{RF}            \\
\midrule
Parameters              & $\mathbf{\sim0.2\,M}$ & $\sim6.7\,M$                 & $\sim6.7\,M$                          & $\sim6.7\,M$            & ---                    \\

MSE                     & $1.36885\times10^{-4}$ & $1.31689\times10^{-4}$      & $\mathbf{1.08245\times10^{-4}}$       & $1.10833\times10^{-4}$  & $1.43304\times10^{-4}$ \\

RMSE                    & $1.16998\times10^{-2}$ & $1.14756\times10^{-2}$      & $\mathbf{1.04041\times10^{-2}}$       & $1.05277\times10^{-2}$  & $1.19710\times10^{-2}$ \\

MAE                     & $4.26589\times10^{-3}$ & $4.19887\times10^{-3}$      & $\mathbf{3.75503\times10^{-3}}$       & $3.76861\times10^{-3}$  & $4.22903\times10^{-3}$ \\

PSNR               & $38.64$                & $38.80$                     & $\mathbf{39.66}$                      & $39.55$                 & $38.44$                \\

Bias                    & $-4.57844\times10^{-5}$ & $-2.00730\times10^{-5}$    & $8.07843\times10^{-5}$                & $6.03971\times10^{-5}$  & $\mathbf{1.67663\times10^{-5}}$ \\

StdError                & $1.16997\times10^{-2}$ & $1.14756\times10^{-2}$      & $\mathbf{1.04037\times10^{-2}}$       & $1.05275\times10^{-2}$  & $1.19710\times10^{-2}$ \\

Global $R^2$            & $0.072$                & $0.108$                     & $\mathbf{0.266}$                      & $0.249$                 & $0.029$                \\

CV error                & $0.0213$               & $0.0209$                    & $\mathbf{0.0189}$                     & $0.0191$                & $0.0218$               \\

SSIM                    & $0.818794$             & $0.826877$                  & $\mathbf{0.851614}$                   & $0.849546$              & $0.799090$             \\
\bottomrule
\end{tabular}
\end{table}

\newpage
\section{Additional Online Results}

\begin{table}[!ht]
\caption{\textbf{Supplementary Table S5 | List of EAMv2 simulations with online ML bias corrections.} 
All simulations use the same configuration described in ~\textbf{Method} section. 
Group~1 corresponds to the free-running control simulation (CLIM), and Group~2 includes EAMv2 runs with four distinct ML correction architectures. 
The correction variables denote the prognostic fields subject to bias correction, and ``ML parameters'' indicate the approximate number of trainable parameters in each model.}
\centering
\scalebox{1.0}{
\begin{tabular}{c l l c}
\hline
\textbf{Group} & \textbf{Simulation} & \textbf{Correction variables} & \textbf{ML parameters} \\
\hline
1 & CLIM       & N/A                & N/A \\
\hline
2 & UNet       & U, V, T, Q         & $\sim$0.2\,M \\
2 & UNetMP     & U, V, T, Q         & $\sim$6.7\,M \\
2 & IUNet      & U, V, T, Q         & $\sim$6.7\,M \\
2 & M\&M       & U, V, T, Q         & $\sim$6.7\,M \\
\hline
\end{tabular}
}
\label{tab:experiments}
\end{table}

\begin{figure}[!ht]
    \centering
    \includegraphics[width=\linewidth]{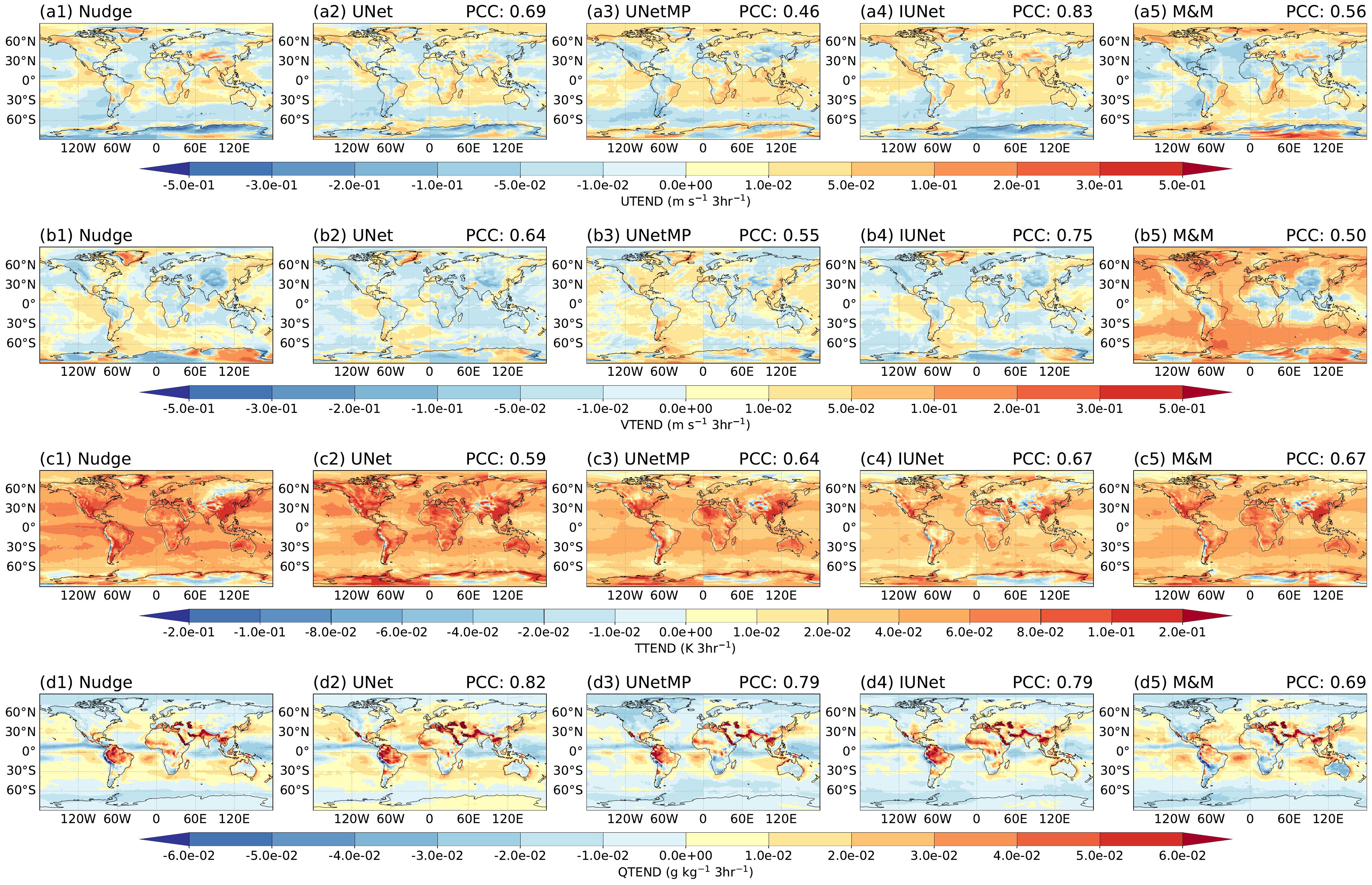}
    \caption{\textbf{Same as main-text Fig.~3, but for the September–October–November (SON) season during 2012–2016.} Shown are the results from EAMv2 simulation nudged towards ERA5 reanalysis (Nudge; panels a1–d1) and four online machine-learning bias-correction simulations: UNet (a2–d2), UNetMP (a3–d3), IUNet (a4–d4), and M\&M (a5–d5). Details of the simulation configurations are provided in Table~\ref{tab:experiments}.}
    \label{fig:son_nudge_tend}
\end{figure}

\begin{figure}[!ht]
    \centering
    \includegraphics[width=\linewidth]{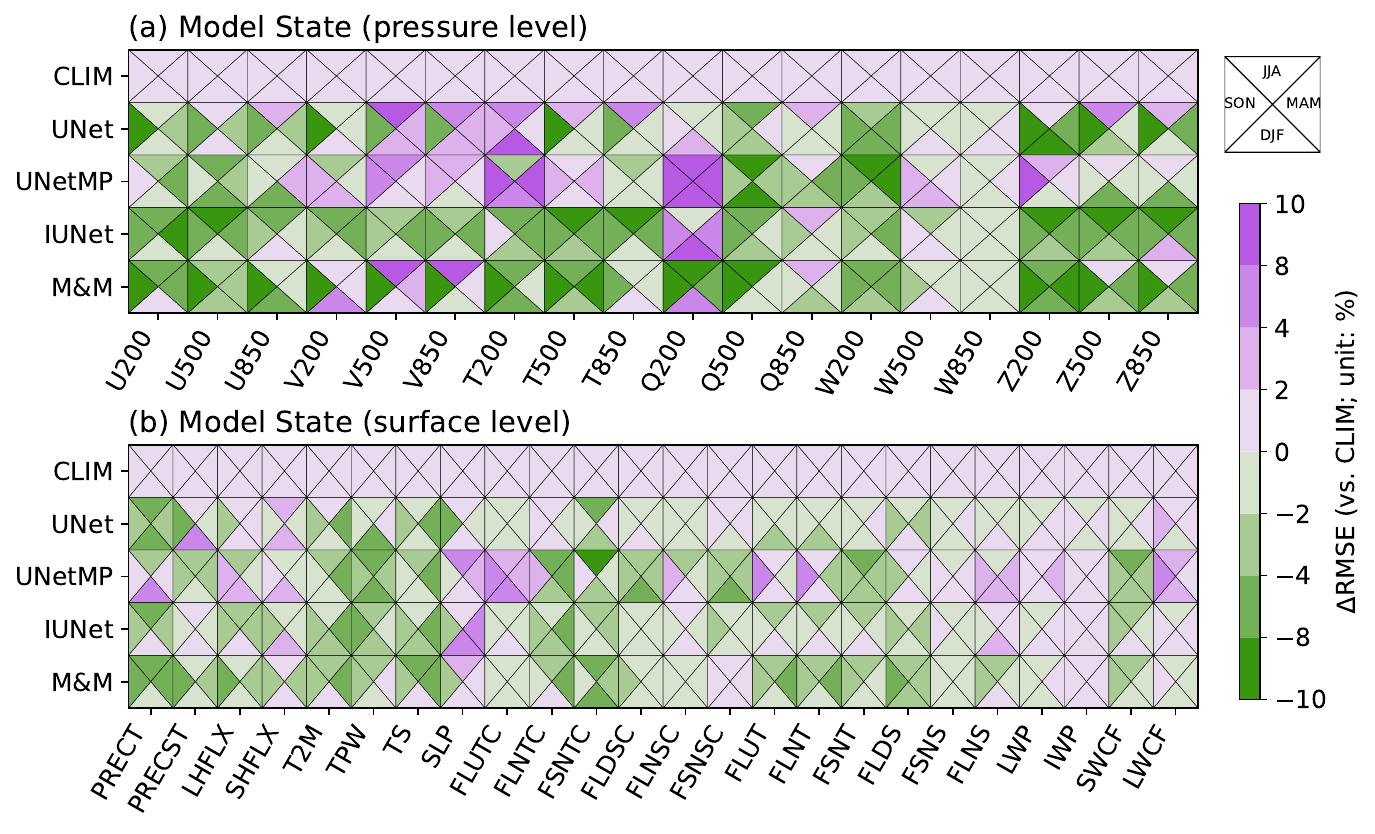}
    \caption{\textbf{Same as main-text Fig.~4, but for 2012–2014.}  The period 2015–2016, which featured a record-breaking El~Niño event, is excluded. Details of the simulation configurations are provided in Table~\ref{tab:experiments}.}
    \label{fig:rmse_heatmap_3yr}
\end{figure}

\begin{figure}[!ht]
    \centering
    \includegraphics[width=\linewidth]{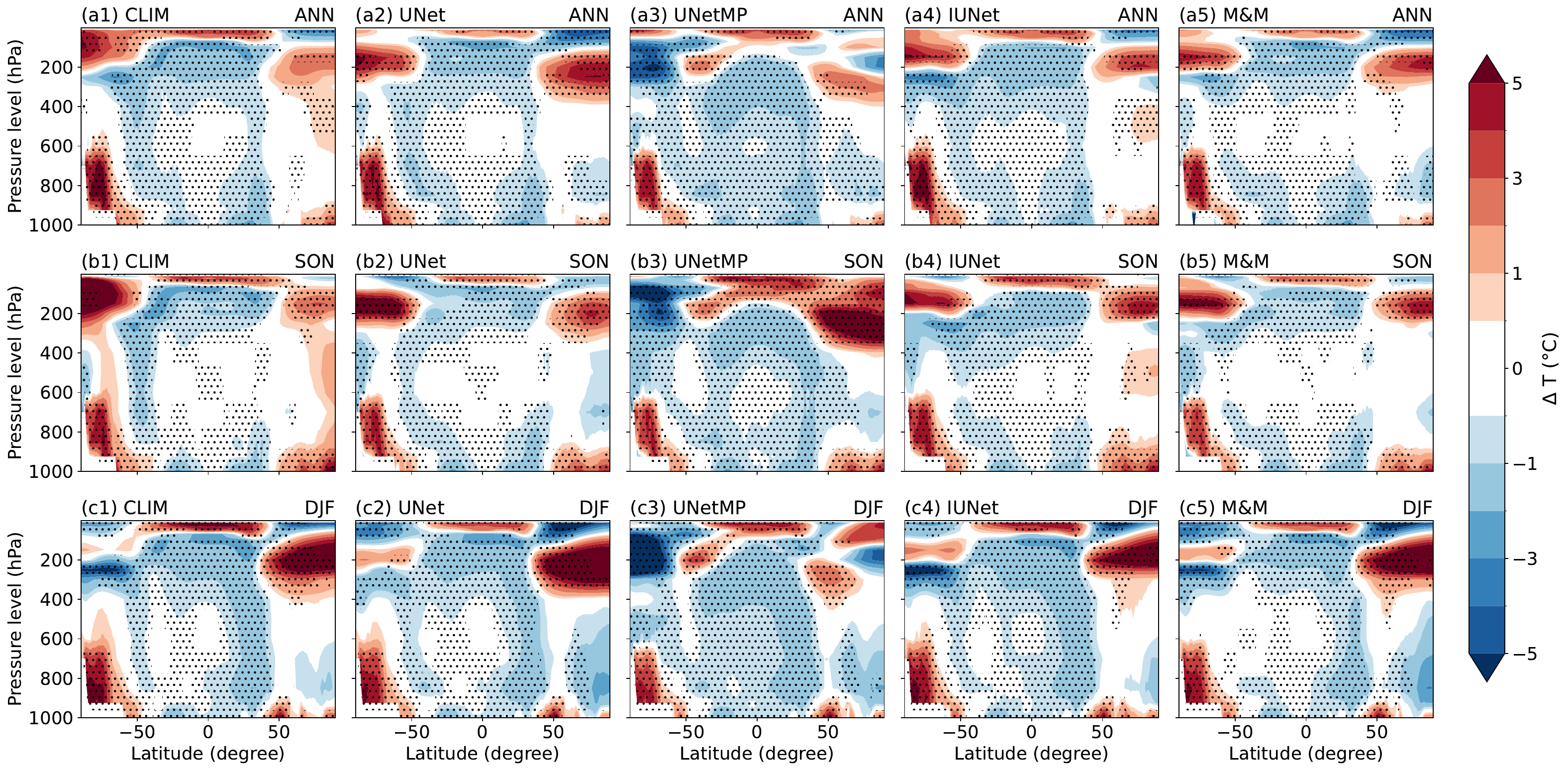}
    \caption{\textbf{Same as main-text Fig.~5, but for zonal-mean temperature biases ($\Delta T$, K).} Shown are the results from the EAMv2 free-running control simulation (CLIM; panels a1–c1) and four online machine-learning bias-correction simulations: UNet (a2–c2), UNetMP (a3–c3), IUNet (a4–d4), and M\&M (a5–c5), compared against ERA5 reanalysis. Details of the simulation configurations are provided in Table~\ref{tab:experiments}.}
    \label{fig:zonal_bias_T}
\end{figure}

\begin{figure}[htbp]
    \centering
    \includegraphics[width=\linewidth]{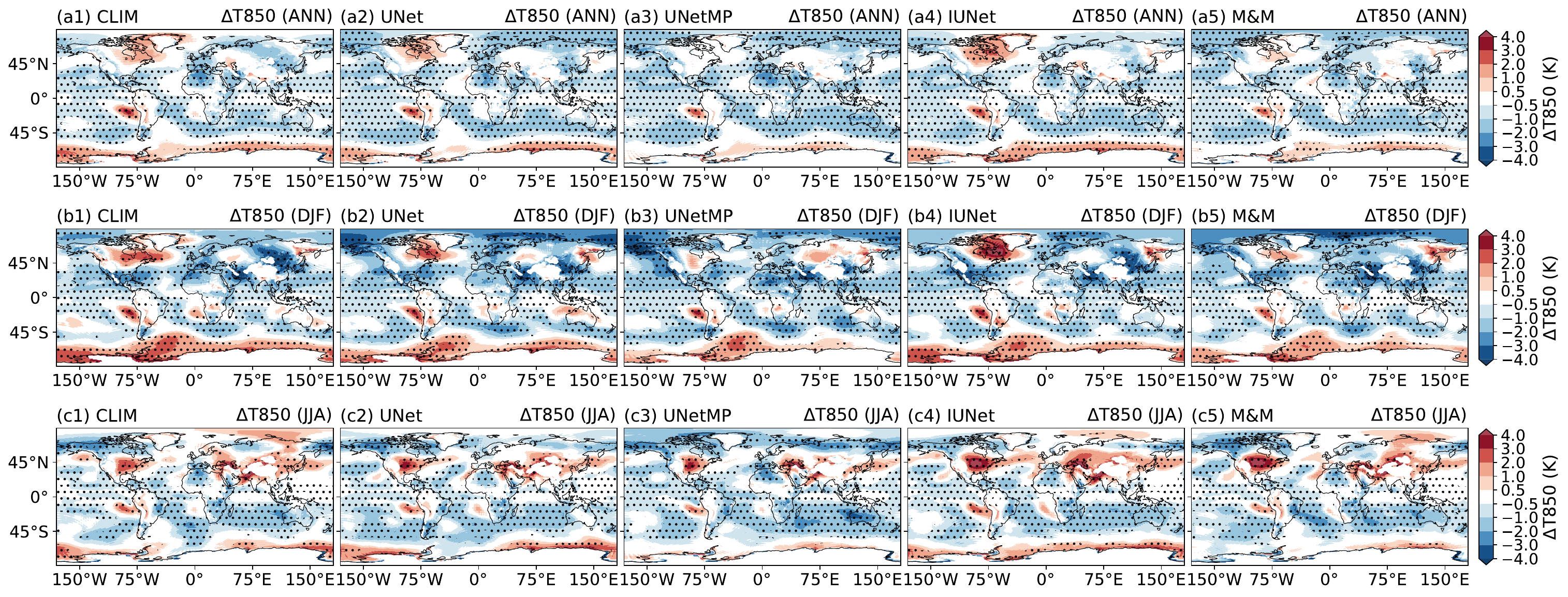}
    \caption{\textbf{Same as main-text Fig.~6, but for 850-hPa temperature biases ($\Delta T_{850}$, K).} 
    Shown are results from the EAMv2 free-running control simulation (CLIM; panels a1–c1) and four online machine-learning bias-correction simulations: UNet (a2–c2), UNetMP (a3–c3), IUNet (a4–d4), and M\&M (a5–c5), compared against ERA5 reanalysis during 2012-2016. Details of the simulation configurations are summarized in Table~\ref{tab:experiments}.}
    \label{fig:temp_bias_map_850hPa}
\end{figure}

\begin{figure}[ht]
\centering
\includegraphics[width=0.65\textwidth]{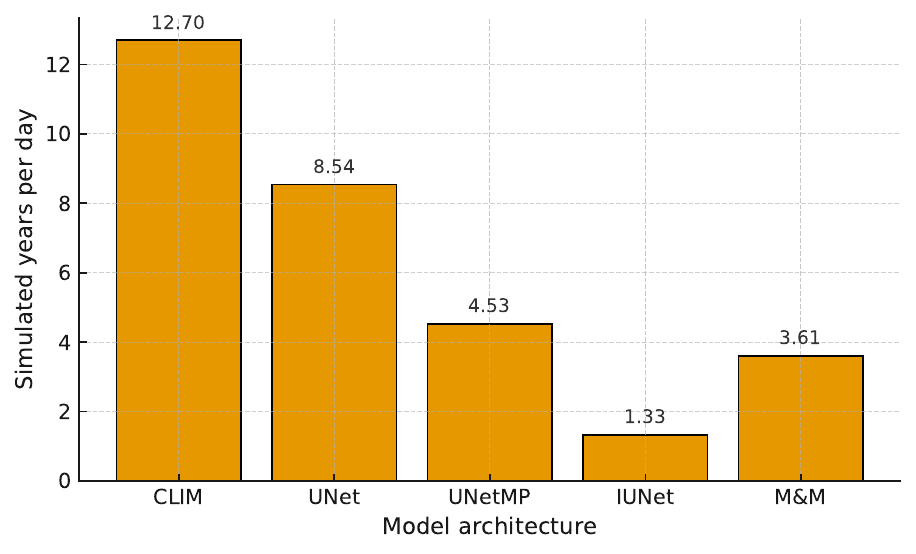}
\caption{\textbf{Computational throughput across model architectures.}
Comparison of simulated years per day (SYPD) achieved by E3SM without (baseline configuration, CLIM) and with machine-learning–based bias-correction architectures (UNet, UNetMP, IUNet, and M\&M) in fully online coupled integrations. All experiments were conducted on CPU nodes of the NERSC Perlmutter system using approximately 1,280 MPI tasks, ensuring consistent hardware configuration across simulations for fair comparison. Details of the experimental setup are provided in Table~\ref{tab:experiments}.}
\label{fig:throughput_comparison}
\end{figure}

\section{Architecture Details}

\begin{figure}[!ht]
    \centering
    \includegraphics[width=0.3\linewidth]{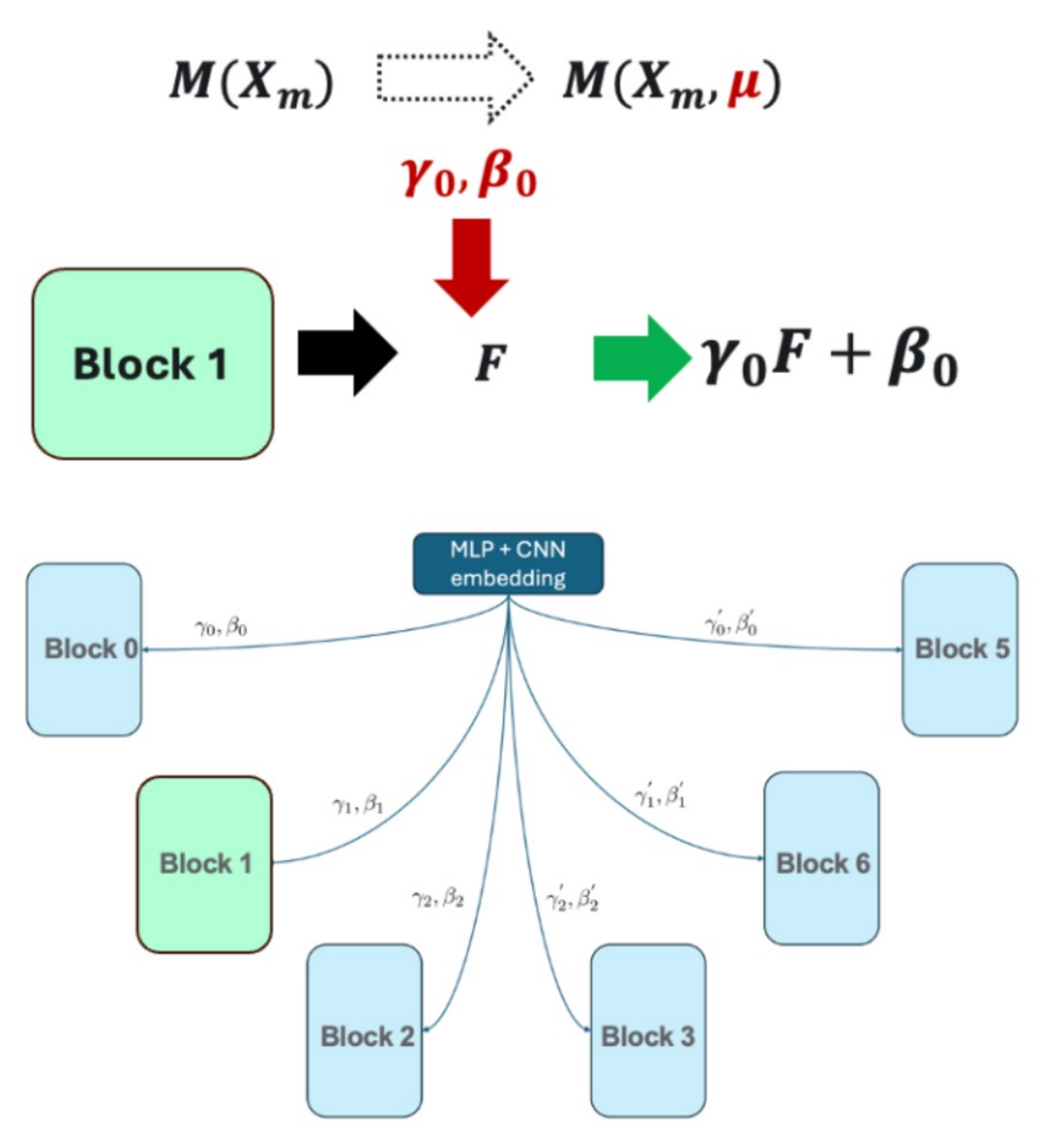}
    \caption{\textbf{FiLM scalar conditioning (level-wise).}
A base model $M(X_m)$ produces features $F$ that are modulated by affine FiLM transforms to yield the conditioned model $M(X_m;\mu)$.  An embedding network (MLP$+$CNN) maps scalar/context metadata $\mu$ to per–resolution-level coefficients $(\gamma_k,\beta_k)$, which are broadcast to each block (Blocks~0–6).  At block $k$, features are transformed as $\mathrm{FiLM}(F_k;\gamma_k,\beta_k)=\gamma_k F_k+\beta_k$ (illustrated for Block~1 with parameters $\gamma_0,\beta_0$).  This level-wise conditioning is applied in all UNet-based architectures evaluated (UNet, IUNet, and M\&M).}
\end{figure}

\begin{figure}[!ht]
    \centering
    \includegraphics[width=\linewidth]{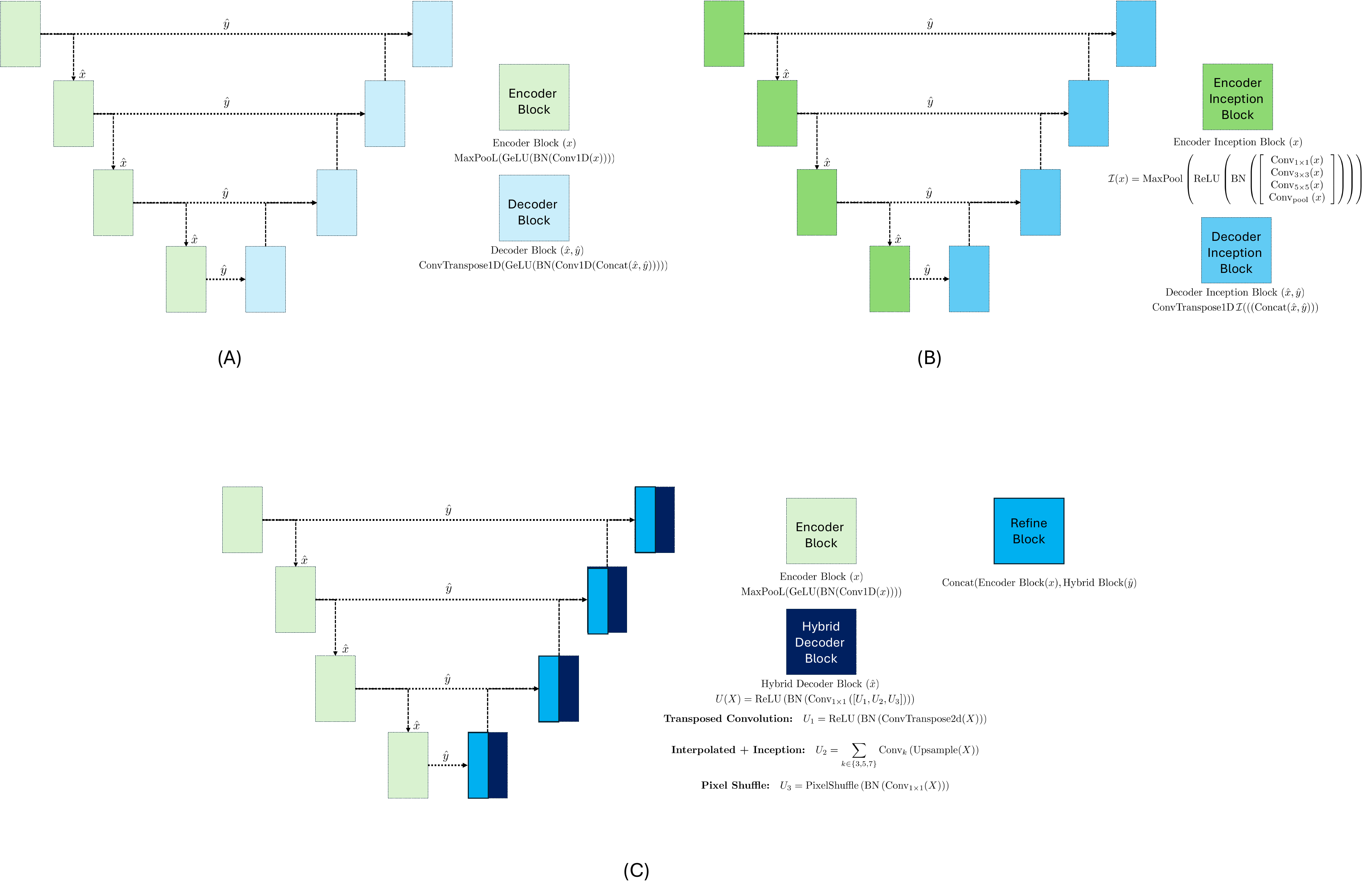}
    \caption{\textbf{Architectures evaluated.}
\textbf{(A) UNet.} Encoder–decoder with symmetric skip connections. Encoder blocks apply Conv1D\,$\rightarrow$\,BatchNorm\,$\rightarrow$\,GeLU with MaxPool downsampling; decoder blocks upsample via ConvTranspose1D, concatenate the corresponding skip features, and refine with a Conv1D layer. \textbf{(B) IUNet.} UNet-style backbone in which each block is an Inception module aggregating parallel $1{\times}1$, $3{\times}3$, and $5{\times}5$ convolutions plus a pooled path; the decoder mirrors this with skip concatenation and ConvTranspose1D upsampling. \textbf{(C) M\&M.} UNet encoder with a hybrid decoder that mixes three upsampling branches: (i) transposed convolution; (ii) interpolation followed by multi-scale convolutions; and (iii) sub-pixel (PixelShuffle). Branch outputs are fused and passed through a Refine block.
Dotted lines denote skip connections; solid arrows indicate forward flow. FiLM scalar conditioning is applied at every resolution level for all three models. For all models trained offline, the input was subsampled to every other latitude–longitude grid point for computational feasibility. During online implementation, the outputs were interpolated back to the original spatial resolution.}
\end{figure}

\end{document}